%% file: WWW/sample-sigconf-authordraft.tex
\documentclass[sigconf]{acmart}
\AtBeginDocument{%
  }

\usepackage{booktabs}
\usepackage{tcolorbox}
\usepackage{caption}
\usepackage{multirow}
\usepackage{subcaption}
\usepackage{setspace}
\usepackage{colortbl}
\usepackage{xcolor}
\usepackage{amsmath}
\usepackage{tabularx}
\usepackage{ragged2e}
\usepackage[table]{xcolor}
\usepackage{pifont}
\usepackage{stfloats}
\usepackage{enumitem}
\usepackage{xcolor}
\definecolor{myblue}{RGB}{0,114,189}
\usepackage{balance}
\copyrightyear{2026}
\acmYear{2026}
\setcopyright{cc}
\setcctype{by}
\acmConference[WWW '26]{Proceedings of the ACM Web Conference 2026}{April 13--17, 2026}{Dubai, United Arab Emirates}
\acmBooktitle{Proceedings of the ACM Web Conference 2026 (WWW '26), April 13--17, 2026, Dubai, United Arab Emirates}
\acmPrice{}
\acmDOI{10.1145/3774904.3793049}
\acmISBN{979-8-4007-2307-0/2026/04}

\begin{document}

\title[TheraMind: A Strategic and Adaptive Agent for Longitudinal
Psychological Counseling]{TheraMind: A Strategic and Adaptive Agent for Longitudinal
Psychological Counseling}
\settopmatter{printacmref=true}

\author{He Hu}
\affiliation{%
  \institution{ Shenzhen University}
  \city{Shenzhen}
  \state{Guangdong}
  \country{China}
}
\email{huhe@gml.ac.cn}

\author{Chiyuan Ma}
\affiliation{%
  \institution{The Chinese University of Hong Kong, Shenzhen}
  \city{Shenzhen}
  \state{Guangdong}
  \country{China}
}
\email{123090409@link.cuhk.edu.cn}

\author{Qianning Wang}
\affiliation{%
  \institution{Auckland University of Technology}
  \city{Auckland}
  \country{New Zealand}
}
\email{yqw7033@autuni.ac.nz}

\author{Lin Liu}
\affiliation{%
  \institution{University of Science and Technology of China}
  \city{Hefei}
  \state{Anhui}
  \country{China}
}
\email{ll0825@mail.ustc.edu.cn}

\author{Yucheng Zhou}
\authornote{Corresponding authors.}
\affiliation{%
  \institution{SKL-IOTSC, CIS, University of Macau}
  \city{Macau}
  \country{China}
}
\email{yucheng.zhou.cs@gmail.com}

\author{Laizhong Cui}
\authornotemark[1]
\affiliation{%
  \institution{Shenzhen University}
  \city{Shenzhen}
  \state{Guangdong}
  \country{China}
}
\email{cuilz@szu.edu.cn}

\author{Fei Ma}
\authornotemark[1]
\affiliation{%
  \institution{Guangdong Laboratory of Artificial Intelligence and Digital Economy (SZ)}
  \city{Shenzhen}
  \state{Guangdong}
  \country{China}
}
\email{mafei@gml.ac.cn}

\author{Qi Tian}
\affiliation{%
  \institution{Guangdong Laboratory of Artificial Intelligence and Digital Economy (SZ)}
  \city{Shenzhen}
  \state{Guangdong}
  \country{China}
}
\email{tian.qi1@huawei.com}

\renewcommand{\shortauthors}{He Hu et al.}

\begin{abstract}
The shortage of mental health professionals has driven the web to become a primary avenue for accessible psychological support. While Large Language Models (LLMs) offer promise for scalable web-based counseling, existing approaches often lack emotional understanding, adaptive strategies, and long-term memory. These limitations pose risks to digital well-being, as disjointed interactions can fail to support vulnerable users effectively. To address these gaps, we introduce TheraMind, a strategic and adaptive agent designed for trustworthy online longitudinal counseling. The cornerstone of TheraMind is a novel dual-loop architecture that decouples the complex counseling process into an Intra-Session Loop for tactical dialogue management and a Cross-Session Loop for strategic therapeutic planning. The Intra-Session Loop perceives the patient's emotional state to dynamically select response strategies while leveraging cross-session memory to ensure continuity. Crucially, the Cross-Session Loop empowers the agent with long-term adaptability by evaluating the efficacy of the applied therapy after each session and adjusting the method for subsequent interactions. We validate our approach in a high-fidelity simulation environment grounded in real clinical cases. Extensive evaluations show that TheraMind outperforms other methods, especially on multi-session metrics like Coherence, Flexibility, and Therapeutic Attunement, validating the effectiveness of its dual-loop design in emulating strategic, adaptive, and longitudinal therapeutic behavior. The code is publicly available at \url{https://github.com/Emo-gml/TheraMind}.

\end{abstract}


\maketitle

\input{WWW/main}

\bibliographystyle{ACM-Reference-Format}
\balance
\bibliography{WWW/sample-base}
\appendix
\input{WWW/appendix}

\end{document}

%% file: WWW/main.tex
\section{Introduction}
The escalating global demand for mental health services has starkly outpaced the availability of trained human therapists, creating a critical gap in care that web-based platforms are increasingly expected to fill \cite{scoping,Survey1}. In response, Large Language Models (LLMs) have emerged as a promising frontier for delivering scalable, accessible, and immediate psychological support to diverse online populations \cite{smile,chatcounselor,catus}. Early explorations have demonstrated their remarkable capacity to generate empathetic and contextually relevant responses, establishing a foundational capability for web applications designed to promote digital well-being \cite{ESCoT,ELIZA}. These initial successes have catalyzed a wave of research aimed at transforming conversational AI from simple chatbots into effective therapeutic partners for online mental healthcare \cite{PsyDial,PsyDT}.

\begin{figure}[!t]
    \centering
    \includegraphics[width=\linewidth]{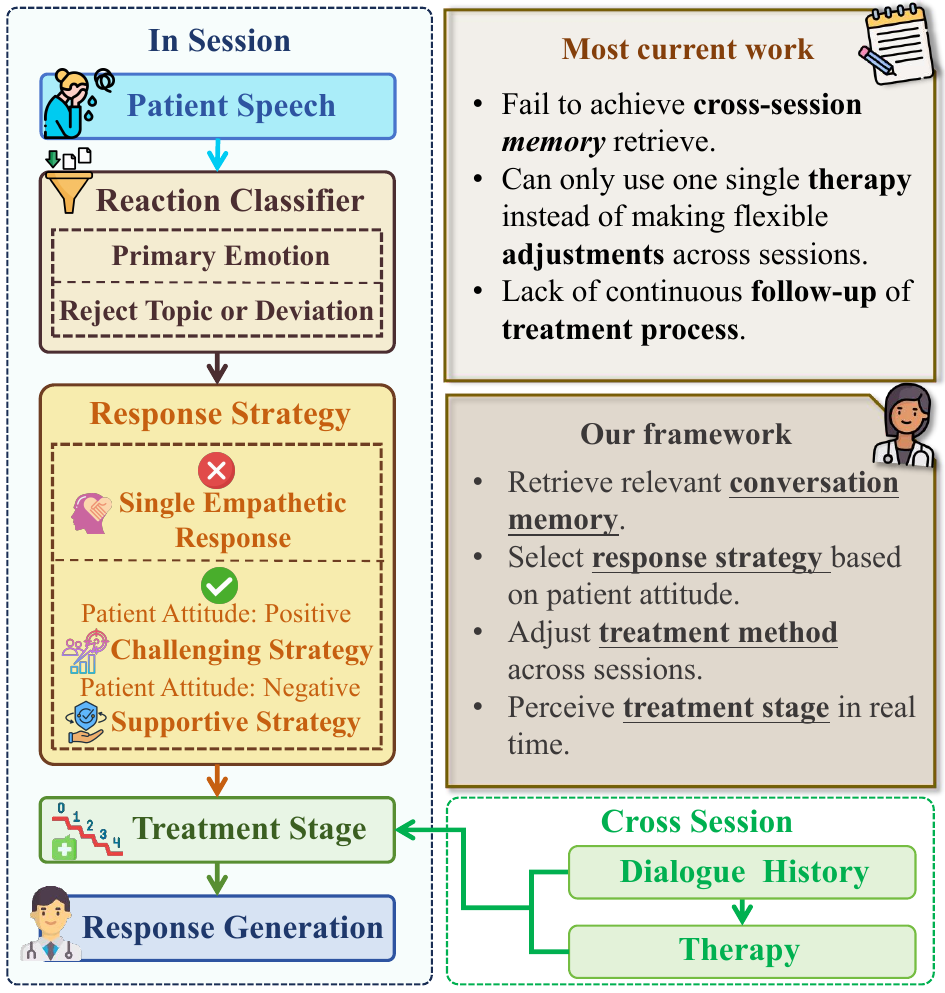}
    \vspace{-5mm}
    \caption{\small Illustration of Our TheraMind Framework.}
    \label{fig:introduction}
    \vspace{-5mm}
\end{figure}

Despite these advancements, current LLM-based counseling agents, such as ChatCounselor and PsyLLM \cite{chatcounselor, Beyond}, operate under a paradigm that fundamentally misaligns with the principles of genuine psychotherapy. These systems are predominantly architected for single-session contexts, exhibiting a form of 'clinical amnesia' that prevents them from building upon past interactions. As illustrated in Figure~\ref{fig:introduction}, they lack the crucial ability to perform \textit{cross-session memory retrieval}, a deficit that erodes the therapeutic alliance and hinders long-term progress. Furthermore, they suffer from strategic rigidity; most are hard-coded with a single therapeutic modality (e.g., Cognitive Behavioral Therapy) and cannot flexibly adjust treatment methods across sessions in response to a patient's evolving needs \cite{treatment,comai2025moving,zarbo2016integrative}. This lack of longitudinal memory and strategic adaptability reduces the complex therapeutic process to disjointed, tactical exchanges, failing to provide the continuous follow-up that is the hallmark of effective care.

To bridge this critical gap between reactive dialogue and strategic therapy, we introduce \textbf{TheraMind}, a strategic and adaptive agent for longitudinal psychological counseling. As illustrated in Figure~\ref{fig:introduction}, the cornerstone of TheraMind is a novel \textbf{dual-loop architecture} that decouples the complex counseling process into an \textit{Intra-Session Loop} for tactical dialogue management and a \textit{Cross-Session Loop} for strategic therapeutic planning. Within a single session, the Intra-Session Loop moves beyond generating simple empathetic responses. It employs a Reaction Classifier to perceive the patient's emotional state and attitude, which in turn informs the dynamic selection of a response strategy, such as a Supportive or Challenging approach, while continuously tracking the current treatment stage. Crucially, the Cross-Session Loop empowers the agent with long-term strategic capabilities. By analyzing the dialogue history after each session, TheraMind evaluates therapeutic efficacy and adaptively selects or adjusts the treatment method for subsequent interactions, ensuring the entire counseling arc is coherent, personalized, and goal-oriented.

To validate our approach, we conducted extensive experiments within a high-fidelity simulation environment grounded in real clinical cases. Our results demonstrate that TheraMind establishes a new state-of-the-art, significantly outperforming a wide range of baselines, including specialized counseling models, general-purpose LLMs, and other agent-based systems. The performance gains are particularly pronounced in multi-session metrics such as Coherence, Flexibility, and Therapeutic Attunement, directly confirming the efficacy of our dual-loop design. Furthermore, both comprehensive ablation studies and human evaluations corroborate that TheraMind's architecture produces more clinically sound, strategically adaptive, and coherent therapeutic dialogues.

This work makes a definitive shift from static, single-turn models to a dynamic, longitudinal agent that more closely emulates the cognitive processes of a human therapist. The primary contributions of this paper are as follows:
\begin{itemize}[left=0pt]
    \item A novel dual-loop agent framework (TheraMind) that models both turn-level dialogue dynamics and the strategic, multi-session structure of psychological counseling.
    \item An adaptive therapy selection mechanism that evaluates therapeutic efficacy and adjusts treatment strategies across sessions, addressing the limitations of static single-therapy approaches.
    \item A clinically grounded dialogue management system that integrates patient state perception, dynamic strategy selection, and treatment phase awareness to support deliberative therapeutic interventions.

\end{itemize}

\section{Related Work}
Current LLM-based counseling approaches lack
emotional understanding, adaptive strategy use,
and dynamic adjustment across sessions with longterm memory, limiting their clinical realism. Intelligent counseling agents aim to address these gaps
by enabling more adaptive, multi-session therapeutic support.
\subsection{Counseling Dialogue Systems}
Counseling dialogue systems are conversational agents for mental health that emphasize empathy, therapeutic alliance, and safety beyond task-oriented systems \cite{pattern,EmoBench,Survey2}. Early work like ELIZA showed that simple pattern matching could emulate therapist-like responses \cite{ELIZA}, inspiring subsequent NLP methods for emotion detection \cite{MELD}, dialogue act classification \cite{Speaker}, and context-aware therapeutic response generation \cite{ESCoT}. Some systems embed structured techniques such as CBT \cite{CBT,catus} or motivational interviewing \cite{Anno-mi}.
Recent efforts adapt LLMs through supervised fine-tuning (SFT) on counseling data. ChatCounselor \cite{chatcounselor} uses the Psych8k corpus to achieve near-ChatGPT performance on counseling benchmarks. CPsyCoun \cite{CPsyCoun} reconstructs Chinese dialogues via a two-phase framework and provides evaluation datasets. MentaLLaMA \cite{MentaLLaMA} explores instruction tuning for empathy and reasoning, while PsyLLM \cite{Beyond} integrates diagnostic and therapeutic reasoning. KokoroChat \cite{KokoroChat} offers 6,589 Japanese dialogues, improving fine-tuned model performance. PsyDial \cite{PsyDial} builds longitudinal datasets via privacy-preserving RMRR, enabling long-term modeling. PsyDT \cite{PsyDT} introduces a digital twin paradigm with dynamic one-shot learning and GPT-4–guided synthesis.
Beyond SFT, reinforcement learning and preference optimization enhance adaptability. COMPEER \cite{COMPEER} applies RL for controllable empathetic reasoning, and \citet{Preference} introduce PsychoCounsel-Preference, a professional preference dataset for reward and policy learning. However, SFT struggles to unify diverse capabilities, while RL is sensitive to reward design \cite{Closer,Reward}, leading to models that cannot fully adapt to real-world therapeutic complexity \cite{Multi-Agent}.

\balance
\subsection{LLM-driven Autonomous Agents}
LLMs such as ChatGPT and Gemini enable autonomous agents integrating reasoning, planning, memory, and tool use \cite{GPT-4,Gemini}. Memory-augmented, context-augmented, and hierarchical architectures allow perception–action loops for complex tasks \cite{Multi-Agent1,zhou2025mam,zhou2023thread,AgentBench,MemoryMamba,sun2025speed,zhou2024visual}. While mostly focused on general tasks, recent work adapts agents for healthcare and mental health: coupling behavioral sensing with proactive support \cite{Transformin_Mental}, simulating therapeutic roles \cite{MIND}, or structuring psychiatric interviews \cite{MAGI}. Specialized agents have emerged, such as AnnaAgent with Multi-Session memory \cite{AnnaAgent}, EmoAgent focusing on safety \cite{EmoAgent}, Interactive Agents for role-play training \cite{qiu2024interactive}, $\Psi$-Arena for interactive benchmarking \cite{Arena}, SweetieChat for strategy-enhanced emotional support \cite{SweetieChat}, and AutoCBT for CBT-based multi-agent counseling \cite{AutoCBT}.

\section{Methodology}
\label{sec:method}

\begin{figure*}[t]
    \centering
    \includegraphics[height=0.5\textheight, width=\linewidth]{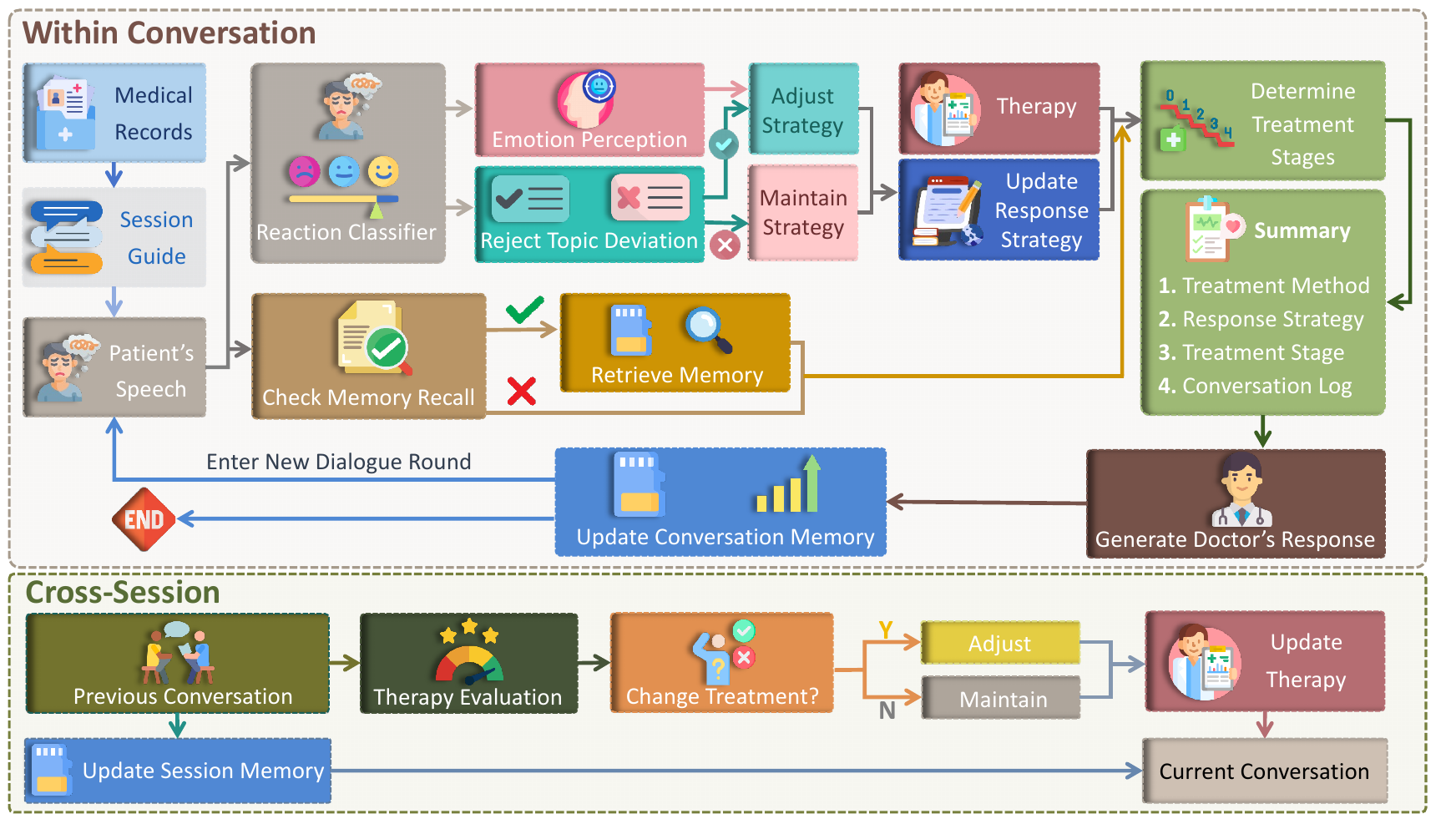}
    \caption{\small TheraMind employs a dual-loop framework. The \textit{Intra-Session Loop} manages turn-level dialogue via state perception, memory retrieval, and clinically grounded responses, while the \textit{Cross-Session Loop} performs session-level planning by evaluating efficacy and adapting strategies, supporting longitudinal and personalized counseling.}
    \label{fig:pipeline}
    \vspace{-4mm}
\end{figure*}

Existing LLM-based counseling agents are predominantly designed as reactive systems, optimized for generating empathetic, single-turn responses within a limited context. This approach fundamentally fails to capture the longitudinal, adaptive, and strategic nature of genuine psychotherapy. To bridge this critical gap, we introduce \textbf{TheraMind}, an autonomous agent framework engineered to emulate the hierarchical cognitive processes of a human therapist over a multi-session treatment arc.

The cornerstone of TheraMind is a novel \textbf{dual-loop architecture}, illustrated in Figure~\ref{fig:pipeline}. This design decouples the complex task of counseling into two distinct yet interconnected control flows:
\begin{enumerate}[left=0pt]
    \item The \textbf{Intra-Session Loop} governs \textit{tactical conversational execution}, managing the nuanced, turn-by-turn dynamics within a single counseling session.
    \item The \textbf{Cross-Session Loop} directs \textit{strategic therapeutic planning}, overseeing the high-level treatment trajectory and its adaptation across multiple sessions.
\end{enumerate}
This hierarchical decomposition enables TheraMind to maintain both immediate conversational coherence and long-term therapeutic objectives simultaneously.

\subsection{The Intra-Session Loop: Dynamic Dialogue Management}
\label{sec:intra_session_loop}
The Intra-Session Loop is the agent's core engine for real-time interaction. Its purpose is to transcend superficial text generation by endowing the agent with perceptual acuity and tactical responsiveness. For each conversational turn $t$, this loop executes a multi-stage process from perception to action.

\subsubsection{Multi-Faceted Patient State Perception}
Human therapists react not just to the literal content of a patient's words, but to the underlying emotional and intentional subtext. Existing models, which primarily process raw text, lack this crucial layer of perception. To address this, we introduce a dedicated perception module.

Upon receiving a patient utterance $U_t$, TheraMind employs a \textbf{Reaction Classifier} to infer a structured representation of the patient's immediate psychological state. Formally, this perception function, $\Phi_{\text{perceive}}$, maps an utterance to a state tuple:
\begin{equation}
    S_t = (E_t, I_t, A_t) = \Phi_{\text{perceive}}(U_t)
\end{equation}
where $E_t$ is the primary \textbf{emotion}, $I_t \in [0, 1]$ is its \textbf{intensity}, and $A_t \in \{\text{Cooperative, Resistant}\}$ is the conversational \textbf{attitude}.
This explicit state representation allows TheraMind to move beyond mere linguistic pattern-matching and engage with the patient's underlying psychological state. It provides a robust, intermediate signal for following decision-making, enabling more nuanced and strategically empathetic interactions.

\subsubsection{Memory-Augmented Contextualization}
Continuity plays a crucial role in shaping a therapist agent's responses during longitudinal psychological counseling. Patients expect therapists to remember significant details from past conversations. Standard LLMs with fixed context windows suffer from catastrophic forgetting, undermining trust and therapeutic progress.

To ensure longitudinal coherence, TheraMind integrates a dynamic memory system. Given the current utterance $U_t$ and the complete history of all previous sessions $\mathcal{H}_{1:k-1}$, a \textbf{Memory Recall} function $\Phi_{\text{memory}}$ produces a relevant memory summary $M_t$:
\begin{equation}
    M_t = \Phi_{\text{memory}}(U_t, \mathcal{H}_{1:k-1})
\end{equation}
This function internally performs a relevance check; if no pertinent history is found, $M_t$ is null.
This mechanism provides a dynamically curated context that is both relevant and succinct. It ensures the agent's responses are historically informed, preventing the disjointedness that plagues stateless models.

\subsubsection{Clinically-Grounded Response Generation}
Effective therapeutic responses are not just empathetic; they are purposeful interventions. Generation must be guided by clear clinical objectives, rather than being an unconstrained generative process. We achieve this by conditioning the LLM on explicit strategic and procedural guidance.

The agent's response, $R_t$, is generated via a deliberative process that integrates multiple streams of information. This process can be formalized as a sequence of steps:
\paragraph{Dynamic Response Strategy.} First, a tactical response strategy $\sigma_t$ is determined from the perceived patient state $S_t$:
\begin{equation}
    \sigma_t = f_{\text{strategy}}(S_t)
\end{equation}
\paragraph{Current Therapeutic Phase.} Concurrently, the current therapeutic phase $P_t$ is identified based on the session's overarching therapy method $\mathcal{T}_k$ and the dialogue history in the current session $H_{t-1}$:
\begin{equation}
    P_t = f_{\text{phase}}(\mathcal{T}_k, H_{t-1})
\end{equation}
\paragraph{Synthesized Generation.} Finally, the agent's response $R_t$ is generated by an LLM $\mathcal{G}$, conditioned on the complete set of factors:
\begin{equation}
    R_t \sim \mathcal{G}( \cdot | U_t, H_{t-1}, \mathcal{T}_k, \sigma_t, P_t, M_t)
\end{equation}
This multi-factor conditioning transforms response generation from a reactive task into a \textbf{deliberative clinical intervention}. Each response is strategically aligned with the patient's immediate state ($S_t$), the conversational tactic ($\sigma_t$), the session's procedural stage ($P_t$), and the long-term therapeutic goal ($\mathcal{T}_k$), ensuring every utterance is purposeful.

\subsection{The Cross-Session Loop: Adaptive Therapeutic Planning}
\label{sec:cross_session_loop}
This loop represents TheraMind's most significant departure from prior work, endowing the agent with the ability to reflect, learn, and adapt its core treatment strategy over time, a hallmark of expert human therapists.

\subsubsection{Post-Session Efficacy Evaluation}
An effective therapist continuously reflects on what works and what does not. To emulate this, an agent must be capable of self-assessment. Without an evaluative step, any multi-session interaction would be a mere sequence of conversations, not a coherent therapeutic process.

At the conclusion of session $k$, with its complete dialogue history $\mathcal{H}_k$, a \textbf{Therapy Evaluation} module assesses the efficacy of the employed therapeutic method $\mathcal{T}_k$. This function yields a quantitative score $\mathcal{E}_k$:
\begin{equation}
    \mathcal{E}_k = f_{\text{eval}}(\mathcal{H}_k, \mathcal{T}_k)
\end{equation}
This reflective step is critical; it transforms the agent from a static system into a learning entity. It provides the necessary feedback signal for long-term strategic adaptation.

\subsubsection{Adaptive Therapy Selection}
Human therapy is inherently adaptive; a single therapeutic modality may not be effective for all patients or for all stages of treatment. Current AI systems are rigid, typically hard-coded to a single therapeutic framework (e.g., CBT). This is a primary barrier to their real-world applicability.

The core innovation of TheraMind resides in its ability to dynamically select the therapeutic method for the subsequent session. Instead of relying on a rigid performance threshold, this decision is delegated to an LLM-based \textbf{Therapy Selection} module, $f_{\text{select}}$. This module performs a nuanced, qualitative assessment, interpreting the quantitative efficacy score $\mathcal{E}_k$ in the context of the session's dialogue history $\mathcal{H}_k$ to make a clinical judgment. The selection process is formalized as:
\begin{equation}
    \mathcal{T}_{k+1} = f_{\text{select}}(\mathcal{T}_k, \mathcal{H}_k, \mathcal{E}_k)
    \label{eq:therapy_update}
\end{equation}
Based on its analysis, this function determines whether the current therapy $\mathcal{T}_k$ should be \textbf{maintained} due to sufficient progress, or if a new, more suitable method should be \textbf{selected} by analyzing the observed outcomes and shortcomings. For instance, if the evaluation highlights a lack of patient engagement with CBT techniques despite a moderate efficacy score, the selection module might reason that a shift to a more client-centered approach is warranted. The initial therapy, $\mathcal{T}_1$, is determined from the patient's intake profile.

This adaptive mechanism is our key contribution to achieving long-term therapeutic effectiveness. It allows TheraMind to personalize the treatment arc, systematically abandoning ineffective strategies and reinforcing successful ones. This capability for \textbf{strategic self-correction}, guided by a deliberative LLM-based judgment, is absent in current counseling agents and is crucial for moving towards truly autonomous therapeutic systems.

\subsection{High-Fidelity Longitudinal Simulation Environment}
Evaluating longitudinal counseling capabilities is notoriously difficult. Static, single-turn datasets are ill-suited for assessing skills like long-term memory, strategic adaptation, and therapeutic alliance building. To overcome this, we propose a new evaluation paradigm grounded in dynamic, high-fidelity simulation.

We construct our simulation environment using real, anonymized clinical cases from the \textbf{CPsyCounR corpus} \cite{CPsyCoun}. To ensure a comprehensive and representative testbed, we sampled 100 cases, selecting 10 from each of the 10 distinct clinical categories present in the dataset, as detailed in Table~\ref{tab:patient_categories}. This stratified sampling approach ensures our agent is evaluated across a wide spectrum of psychological issues.

\begin{table}[!t]
\centering
\caption{\small Distribution of Patient Profile Categories Sampled from the CPsyCounR Corpus.}
\label{tab:patient_categories}
\vspace{-2mm}
\resizebox{\linewidth}{!}{%
\begin{tabular}{ll}
\toprule
\textbf{Category} & \textbf{Description of Psychological Issues} \\
\midrule
Love & Emotional \& Intimacy Problems (e.g., romantic, marital) \\
Family & Family \& Parent-Child Relationship Issues \\
Emotion & Emotional Disorders (e.g., depression, anxiety, OCD) \\
Youth & Child \& Adolescent Psychological/Behavioral Problems \\
Social & Interpersonal Relationships \& Social Adaptation \\
Stress & Stress \& Stress-Related Disorders (e.g., academic, occupational) \\
Addiction & Addiction \& Behavioral Control Issues (e.g., internet, gambling) \\
Anxiety & Neurosis \& Somatization Problems \\
Self-growth & Personality \& Personal Growth Issues (e.g., self-cognition) \\
Rare & Special Psychological Disorders (e.g., schizophrenia, autism) \\
\bottomrule
\end{tabular}%
}
\vspace{-2mm}
\end{table}

For each selected case, an initialization process, $\Phi_{\text{init}}$, generates a structured \textbf{Patient Profile} ($\mathcal{P}$) and a series of high-level \textbf{Session Guides} ($\mathcal{G} = \{g_1, \dots, g_K\}$) for a six-session arc:
\begin{equation}
    (\mathcal{P}, \mathcal{G}) = \Phi_{\text{init}}(\text{Case}_{\text{CPsyCounR}})
\end{equation}
These guides provide the LLM-based patient simulator with coherent goals for each session without rigidly scripting responses. This flexibility enables the simulated patient to exhibit a dynamic emotional range, capable of expressing both positive and negative attitudes, thereby mirroring the variability of real-world patients. This semi-structured approach creates a controllable yet dynamic testbed that balances clinical realism with reproducibility, enabling a robust evaluation of advanced, multi-session counseling skills.

\section{Experiments}

\subsection{Compared Methods}
To comprehensively evaluate the performance of TheraMind, we evaluate it with other methods, which we categorize into three groups: (1) \textbf{Psychological counseling models}: ChatCounselor \cite{chatcounselor}, CPsyCounX \cite{CPsyCoun}, and PsyDTLLM \cite{PsyDT}; (2) \textbf{General-purpose large language models}: GLM-4-9B-Chat \cite{glm2024chatglm}, InterLM2.5-7B-Chat \cite{cai2024internlm2}, Qwen3-8B \cite{yang2025qwen3}, Kimi-Dev-72B \cite{team2025kimi}, Doubao-1.5-pro-32k \cite{guo2025seed1}, Yi-Large \cite{young2024yi}, and DeepSeek-V3 \cite{liu2024deepseek}; and (3)\textbf{Psychological agent-based models}: Interactive Agents \cite{qiu2024interactive}.

\subsection{Evaluation Dataset and Metrics}

\subsubsection{Evaluation Dataset}
Our experiments use the \textbf{CPsyCounR} dataset \cite{CPsyCoun}, containing 3,134 anonymized and professionally rewritten Chinese psychological counseling reports from Yidianling \cite{yidianling} and Psy525 \cite{psy525}. Each report follows a standardized structure with a case brief, consultation process, and therapist reflections, ensuring authenticity, privacy, and clinical reliability, and supporting realistic evaluation of long-term counseling agents.

\subsubsection{Evaluation Metrics}
To provide a comprehensive and clinically grounded assessment, our evaluation protocol draws upon established psychotherapy research and professional counseling standards \cite{hill1999helping,Psychology,PsyDial,CPsyCoun}. It is structured into single-session and multi-session metrics. Trained annotators rate all metrics on a 4-point scale (0-3).

\textbf{Single-Session Metrics.} We evaluate the quality of individual interactions along two primary dimensions:
\begin{enumerate}[leftmargin=*]
    \item \textbf{Therapeutic Alliance (T.Alli):} Based on the Working Alliance Inventory (WAI) framework \cite{working}, this metric evaluates the agent's ability to form a collaborative partnership. It measures alignment on therapeutic \textit{Goals}, collaborative execution of \textit{Tasks}, and the establishment of an empathetic \textit{Bond}.
    \item \textbf{Interaction Quality (Inter):} This metric measures clinical dialogue depth by assessing the agent’s ability to integrate patient motivations, probe cognitive contradictions, and foster self-awareness.
    
\end{enumerate}

\textbf{Multi-Session Metrics.} To evaluate longitudinal performance, we introduce four metrics that capture the agent's effectiveness over the entire therapeutic arc:
\begin{enumerate}[leftmargin=*]
    \item \textbf{Coherence (Coh):} Measures the agent's ability to maintain continuity across sessions by referencing and building upon past conversations to create a cohesive therapeutic narrative.
    \item \textbf{Flexibility (Flex):} Assesses the capacity to adapt therapeutic strategies over time, adjusting or integrating different methods in response to the patient's evolving needs and progress.
    \item \textbf{Empathy (Emp):} Evaluates the consistency of the agent's empathetic engagement, including its ability to track emotional development, reframe negative expressions constructively, and utilize patient-generated metaphors as therapeutic tools.
    \item \textbf{Therapeutic Attunement (T.Attun):} Measures how well the agent aligns interventions with the treatment stage, integrates techniques, and detects subtle therapeutic progress.

\end{enumerate}

\begin{table*}[!t]\small
\centering
\caption{Performance comparison on single-session and multi-session psychological counseling metrics. Scores are on a 4-point scale (0-3), with higher values being better.}
\label{tab:main1}
\vspace{-3mm}
\setlength{\tabcolsep}{5.5mm}
\begin{tabular}{@{}lccccccccc@{}}
\toprule
\multirow{2}{*}{\textbf{Model}} & \multicolumn{3}{c}{\textbf{Single-Session}} & \multicolumn{5}{c}{\textbf{Multi-Session}} \\
\cmidrule(lr){2-4} \cmidrule(lr){5-9}
 & T.Alli & Inter & \textbf{Avg} & Coh & Flex & Emp & T.Attun & \textbf{Avg} \\
\midrule
ChatCounselor \cite{chatcounselor} & 0.237 & 0.038 & 0.138 & 0.472 & 0.541 & 0.384 & 0.363 & 0.440 \\
CPsyCounX \cite{CPsyCoun} & 0.374 & 0.063 & 0.219 & 0.640 & 0.710 & 0.550 & 0.440 & 0.585 \\
GLM-4-9B-Chat \cite{glm2024chatglm} & 0.571 & 0.150 & 0.361 & 1.150 & 1.560 & 1.380 & 1.360 & 1.363 \\
InterLM2.5-7B-Chat \cite{cai2024internlm2} & 0.644 & 0.198 & 0.421 & 1.210 & 1.570 & 1.670 & 1.720 & 1.543 \\
Qwen3-8B \cite{yang2025qwen3} & 1.035 & 0.598 & 0.817 & 1.440 & 1.610 & 2.150 & 2.000 & 1.800 \\
Kimi-Dev-72B \cite{team2025kimi} & 0.837 & 0.518 & 0.678 & 1.590 & 1.810 & 1.850 & 1.970 & 1.805 \\
Doubao-1.5-pro-32k \cite{guo2025seed1} & 0.643 & 0.497 & 0.570 & 1.510 & 1.980 & 1.950 & 1.830 & 1.818 \\
Yi-Large \cite{young2024yi} & 0.860 & 0.485 & 0.673 & 1.650 & 1.860 & 2.080 & 2.070 & 1.915 \\
PsyDTLLM \cite{PsyDT} & 0.963 & 0.526 & 0.745 & 1.660 & 1.760 & 2.240 & 2.090 & 1.938 \\
DeepSeek-V3 \cite{liu2024deepseek} & 1.978 & 1.712 & 1.845 & 2.160 & 2.100 & 2.590 & 2.470 & 2.330 \\
Interactive Agents \cite{qiu2024interactive} & 1.424 & 2.373 & 1.899 & 2.390 & 2.060 & 2.570 & 2.320 & 2.335 \\
\midrule
\textbf{TheraMind} & \textbf{2.210} & \textbf{2.505} & \textbf{2.358} & \textbf{2.860} & \textbf{2.290} & \textbf{2.980} & \textbf{2.890} & \textbf{2.755} \\
\bottomrule
\end{tabular}
\vspace{-1mm}
\end{table*}

\subsection{Experimental Settings}

We select DeepSeek-V3 \cite{liu2024deepseek} as the backbone model, owing to its cost efficiency and competitive performance on instruction-following tasks. For response generation, the agent is configured with a temperature of $0.9$, $top\text{-}p = 0.75$, and $top\text{-}k = 20$. In judgment tasks, we adopt a lower temperature of $0.3$ while keeping $top\text{-}p$ and $top\text{-}k$ unchanged, to encourage more deterministic and stable outputs.

Unlike previous research that relies on predefined datasets, which suffer from two major drawbacks: (1) patient responses tend to be overly simplistic, failing to capture the diversity of real-world cases; (2) replies are fixed and lack cross-turn consistency, we ground our evaluation in the CPsyCounR dataset \cite{CPsyCoun}. It follows a standardized structure including \emph{Title}, \emph{Type}, \emph{Method}, \emph{Case Brief}, \emph{Consultation Process}, and \emph{Experience Thoughts}. To ensure representativeness, we uniformly sampled 100 cases across various patient categories and conducted evaluations over six-session dialogues. Based on these real cases, we employed a large language model to simulate virtual patients capable of flexibly displaying both positive and negative attitudes, thereby building a more realistic and diverse evaluation.

\begin{figure}[!t]
    \centering
    \includegraphics[width=0.9\linewidth,trim=0 5 0 0,clip]{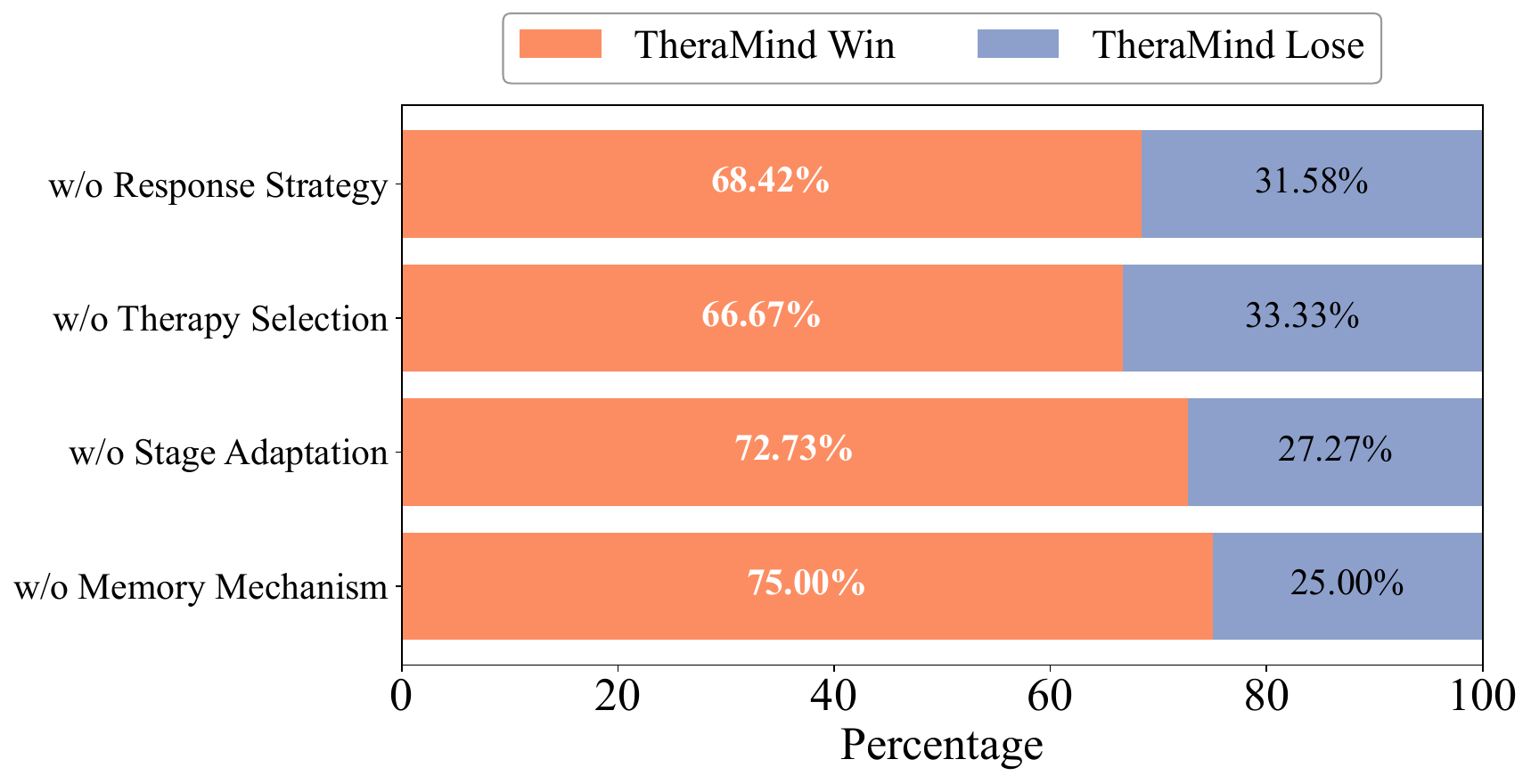}
    \caption{\small Ablation Study. Pairwise comparison between the full TheraMind model and its ablated variants.}
    \label{fig:ablation}
    \vspace{-3mm}
\end{figure}

\subsection{Main Results}
As shown in Table~\ref{tab:main1}, TheraMind establishes a new state-of-the-art by significantly outperforming all baseline methods across both single-session and multi-session metrics.
While TheraMind achieves the highest single-session scores (\textbf{2.358} Avg), its most substantial performance gains are in the multi-session evaluation, with leading scores in Coherence (\textbf{2.860}), Empathy (\textbf{2.980}), and Therapeutic Attunement (\textbf{2.890}). This highlights the effectiveness of the Intra-Session Loop's mechanisms for maintaining long-term, clinically-grounded interactions. Crucially, its superior score in Flexibility (\textbf{2.290}) directly confirms the efficacy of the Cross-Session Loop's adaptive therapy selection, a capability largely absent in prior work.
Furthermore, the comparison with its backbone model, DeepSeek-V3, is telling. The TheraMind agent framework elevates performance dramatically across all metrics, boosting the multi-session average from 2.330 to \textbf{2.755} (a 18.2\% relative improvement). This demonstrates that the agent architecture provides critical strategic and adaptive reasoning capabilities far beyond the generative function of the base LLM.

\begin{figure*}[!t]
    \centering
    \includegraphics[width=0.85\linewidth]{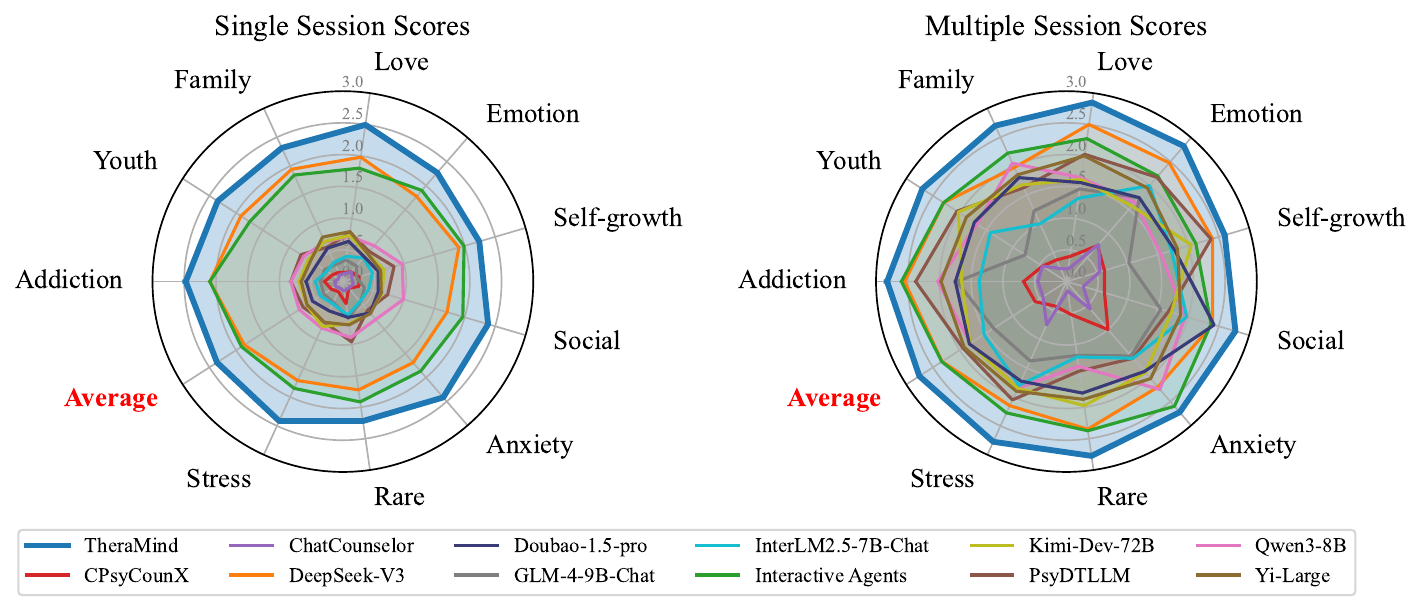}
    \vspace{-4mm}
    \caption{\small Performance comparison across ten distinct categories of psychological issues, evaluated in both single-session (left) and multi-session (right) contexts. The axes correspond to the categories detailed in Table~\ref{tab:patient_categories}.}
    \label{fig:radar}
    \vspace{-2mm}
\end{figure*}

\begin{figure*}[t]
    \centering
    \includegraphics[width=0.29\linewidth]{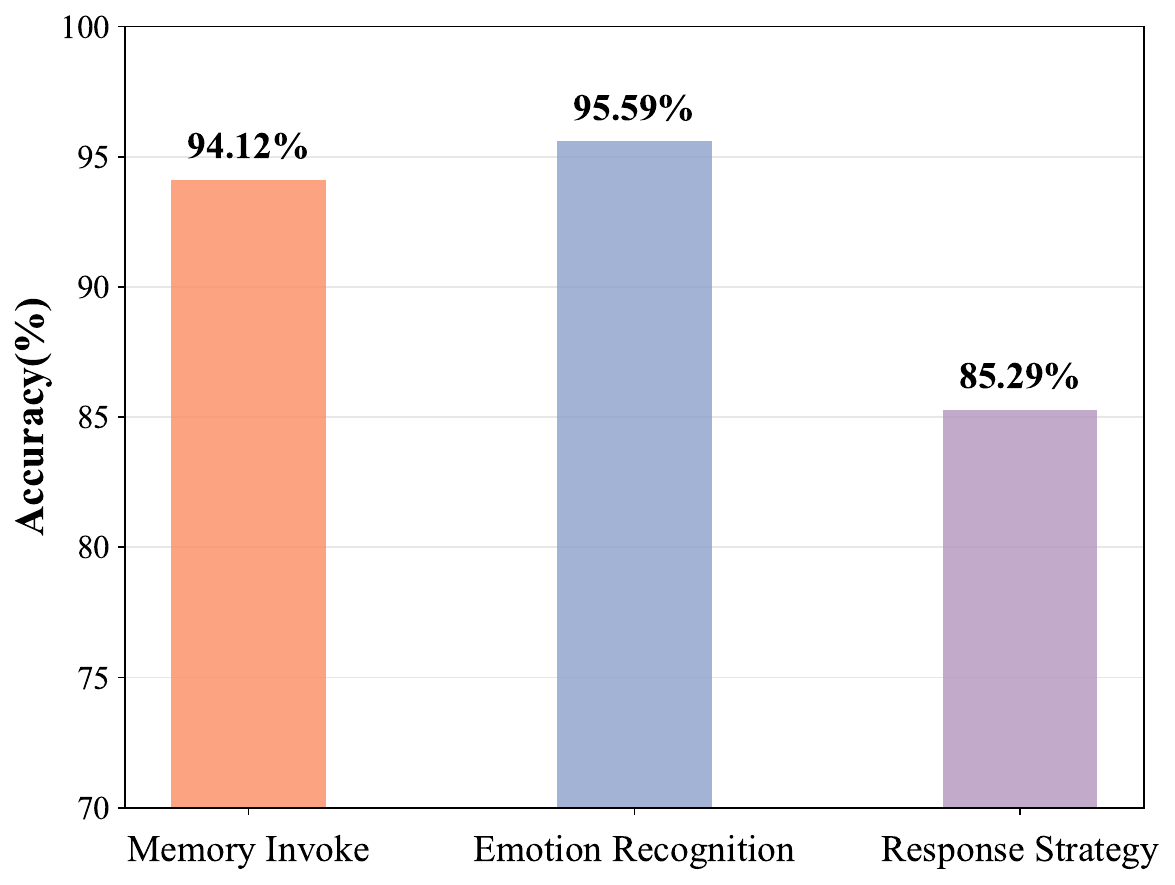}\hfill
    \includegraphics[width=0.38\linewidth]{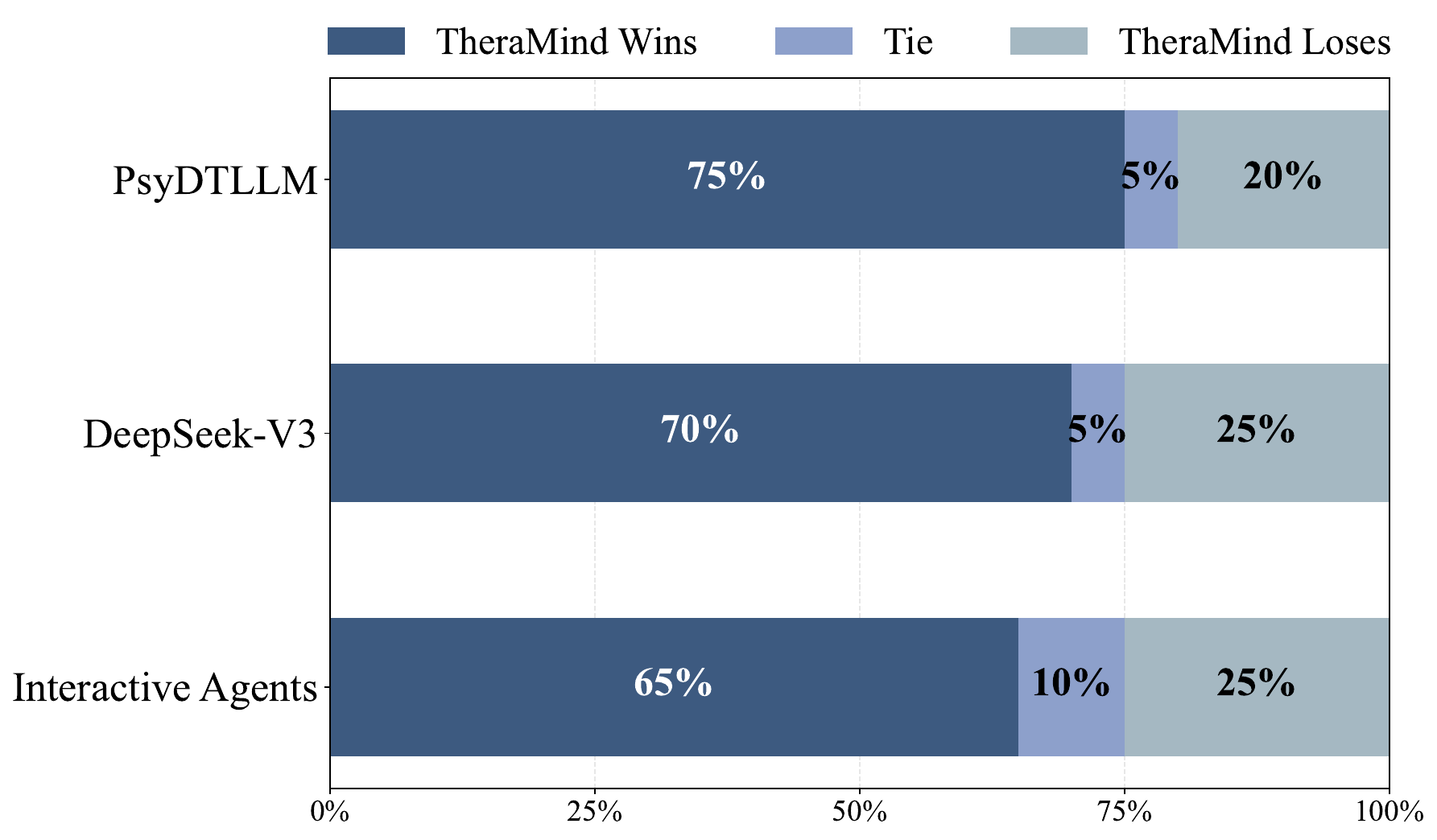}\hfill
    \includegraphics[width=0.27\linewidth]{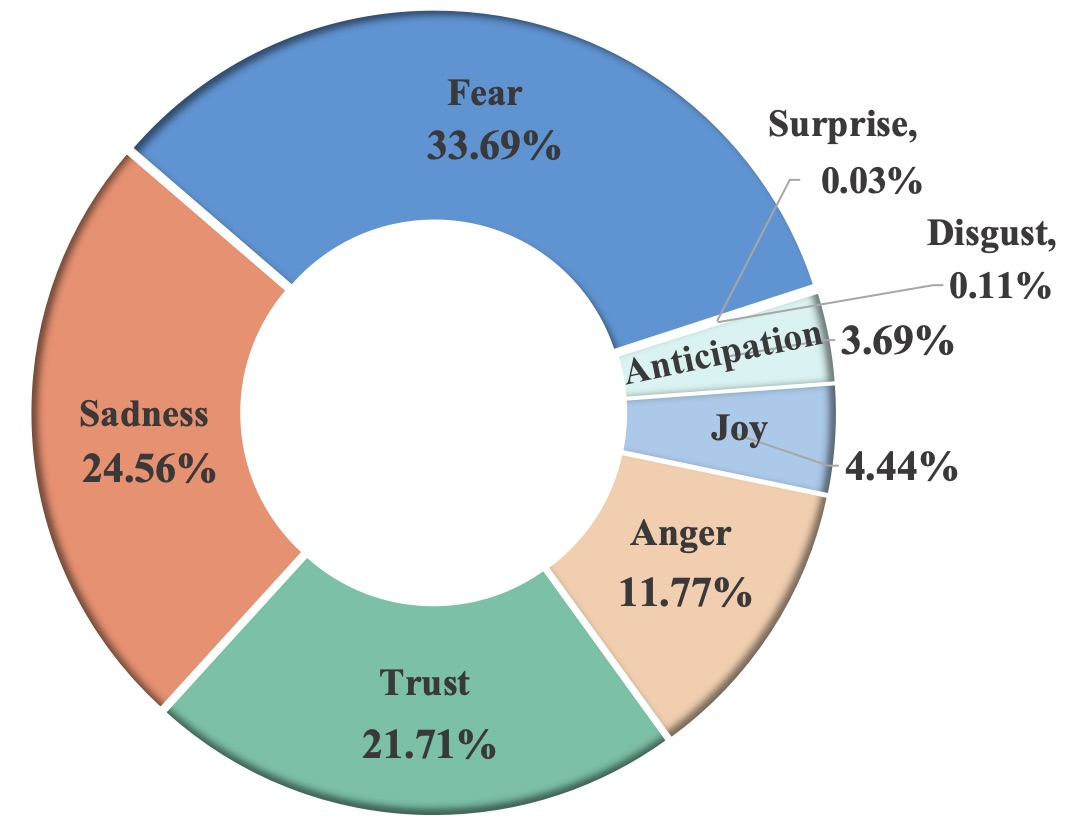}
    \vspace{-3mm}
    \caption{\small (Left) Human agreement on TheraMind’s internal decisions. (Middle) Pairwise preference comparison with three strong baselines in a simplified multi-session setting. (Right) Emotion distribution identified by TheraMind during counseling.
}
    \label{fig:agreement}
    \label{fig:comparison}
    \label{fig:bing}
    \vspace{-3mm}
\end{figure*}

\subsection{Ablation Study}
To validate the contribution of each core component within TheraMind's dual-loop architecture, we conducted an ablation study via pairwise comparison. We evaluated the full model against variants lacking specific modules: Memory Mechanism, Stage Adaptation, Response Strategy, and Therapy Selection. Figure~\ref{fig:ablation} presents the win rates of the full TheraMind model against these ablated versions.

The results clearly show the necessity of all components, with the full model outperforming all ablated variants. Removing the \textbf{Memory Mechanism} causes the largest drop (75.00\% win rate), underscoring the importance of maintaining longitudinal context to avoid ``clinical amnesia''. Excluding \textbf{Stage Adaptation} (72.73\%) highlights the need to align responses with therapeutic phases. Removing \textbf{Response Strategy} (68.42\%) and \textbf{Therapy Selection} (66.67\%) increases rigidity within and across sessions, confirming the value of adaptive, multi-granular control.

\subsection{Analysis Across Psychological Issues}

To assess the robustness and generalizability of TheraMind, we conducted a fine-grained performance analysis across the ten distinct categories of psychological issues. The results, as shown in Figure~\ref{fig:radar}, reveal critical insights into the models' capabilities in both single-turn and longitudinal contexts. In the single-session evaluation (left panel), TheraMind consistently achieves the highest scores across all domains. Moreover, the multi-session evaluation (right panel) paints a much clearer picture of TheraMind's unique advantages. The performance gap between TheraMind and all other models widens dramatically. While most models show some improvement, their gains are inconsistent and marginal compared to TheraMind, which maintains a consistently high level of performance across all ten categories. This demonstrates the profound impact of our dual-loop architecture. Lacking a mechanism for cross-session memory and strategic adaptation, competing models fail to build a coherent, long-term therapeutic narrative, causing their performance to plateau or degrade over time. In contrast, TheraMind's ability to retrieve memories and adapt its therapy method allows it to effectively handle complex, evolving issues, showcasing its superior capability for true longitudinal counseling.

\subsection{Human Evaluation}
To complement our automatic metrics and validate the nuanced clinical capabilities of our agent, we conducted two sets of human evaluations. Trained annotators with expertise in psychology performed all assessments.

First, we assessed the internal validity of TheraMind's core components by evaluating the agreement between the agent's decisions and human judgment. On a sample of dialogues from 10 distinct cases, we achieved a Cohen's~$\kappa$ \cite{cohens} of 0.676, indicating substantial inter-annotator reliability, which falls within the range of substantial agreement (0.60–0.80). As shown in Figure~\ref{fig:agreement} (Left), TheraMind's modules demonstrate high fidelity with human clinical reasoning. The \textbf{Emotion Recognition} (95.59\% agreement) and \textbf{Memory Invoke} (94.12\% agreement) modules are exceptionally accurate, confirming the robustness of the agent's perceptual and contextualization capabilities. The \textbf{Response Strategy} selection also shows strong alignment (85.29\%), validating its ability to make clinically sound tactical decisions.

Second, we conducted a pairwise comparative evaluation of TheraMind against three top-performing baselines: PsyDTLLM, DeepSeek-V3, and Interactive Agents. For this, we used a simplified multi-session evaluation standard on 20 cases (two from each category), achieving a Cohen's~$\kappa$ of 0.697. The results, presented in Figure~\ref{fig:comparison} (Middle), reveal a clear human preference for TheraMind. It secured decisive win rates against the specialized counseling model PsyDTLLM (75\%), its powerful backbone model DeepSeek-V3 (70\%), and the advanced agent framework Interactive Agents (65\%). This human-validated superiority demonstrates that our dual-loop architecture is the key factor in achieving a higher standard of longitudinal therapeutic care.

\subsection{In-depth Analysis of TheraMind's Behavior}

To gain deeper insights into TheraMind’s therapeutic process and internal decision-making, we conduct a multi-faceted analysis of its behavior in simulated counseling sessions, covering emotional dynamics, intervention strategies, therapeutic phase progression, and the relationship between patient attitude and emotional intensity.

As illustrated in Figure \ref{fig:bing} (Right), the emotional landscape of the interactions is dominated by expected therapeutic emotions such as Fear (33.69\%) and Sadness (24.56\%). Crucially, Trust emerges as the third most prevalent emotion at 21.71\%. This high proportion of Trust is a strong indicator of the agent's success in establishing a robust therapeutic alliance, which encourages deeper patient self-disclosure.

\begin{figure*}[!t]
    \centering
    \includegraphics[width=0.38\linewidth]{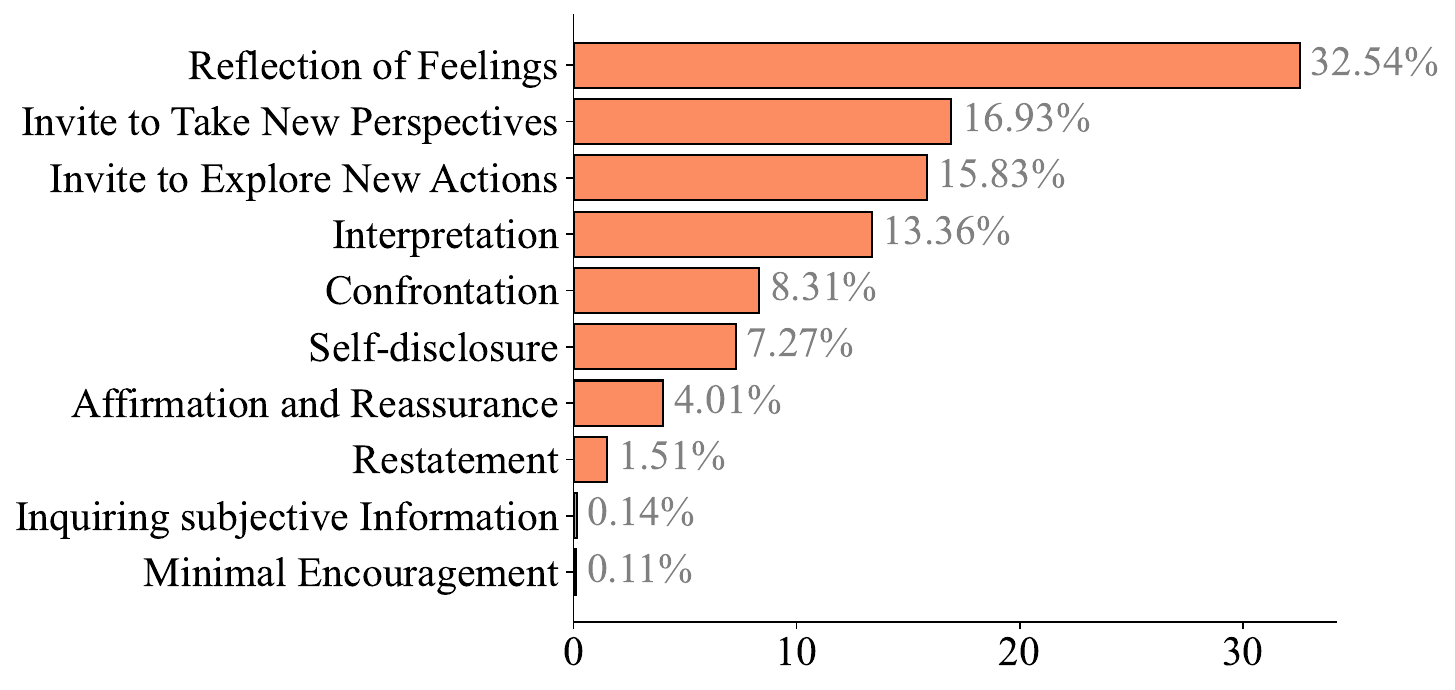}
    \includegraphics[width=0.31\linewidth]{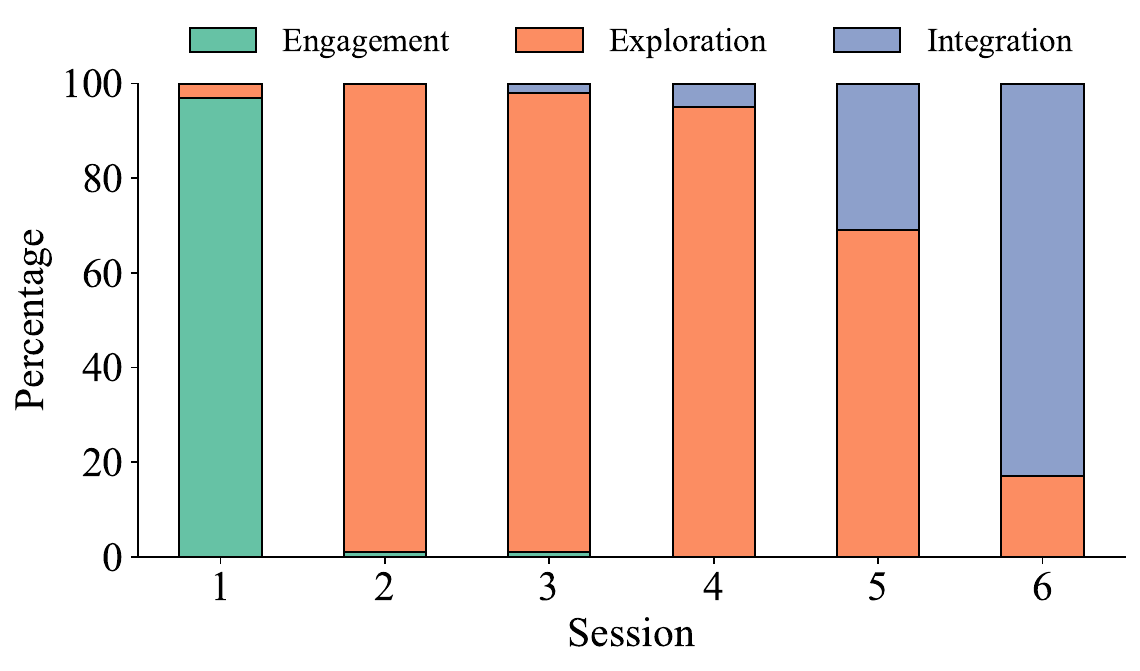}
    \includegraphics[width=0.29\linewidth]{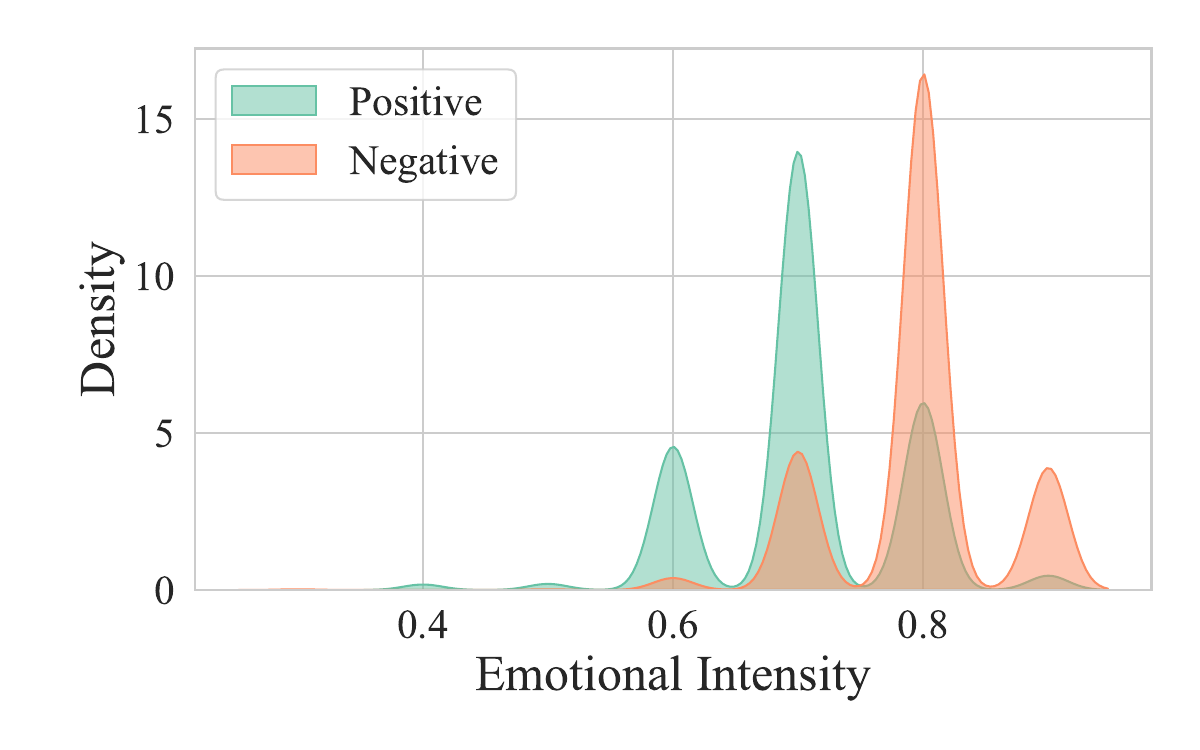}
    \vspace{-3mm}
    \caption{\small (Left) Distribution of TheraMind’s intervention strategies. (Middle) Session-wise distribution of counseling phases. (Right) KDE of emotional intensity by patient attitude (positive vs. negative).
}
    \label{fig:output}
    \label{fig:phase}
    \label{fig:attitude}
    \vspace{-3mm}
\end{figure*}

An analysis of the agent's intervention strategies, shown in Figure \ref{fig:output} (Left), reveals a clinically-grounded and diverse therapeutic repertoire. The most frequently used intervention is ``Reflection of Feelings'' (32.54\%), underscoring the agent's focus on building empathy. However, TheraMind does not rely solely on supportive responses. It actively guides the therapeutic process by employing change-oriented techniques such as ``Invite to Take New Perspectives'' (16.93\%) and ``Invite to Explore New Actions'' (15.83\%), demonstrating a balanced approach that facilitates both emotional processing and cognitive-behavioral change.

Furthermore, TheraMind captures the structured progression of a longitudinal therapeutic arc. Figure~\ref{fig:phase} (Middle) shows a clear transition across clinical phases over six sessions: early \textit{Engagement}, mid-stage \textit{Exploration}, and late-stage \textit{Integration}. This ability to guide counseling through a coherent multi-stage process distinguishes TheraMind from static, single-session models.

Finally, our framework captures nuanced links between patient attitude and emotional expression, as shown in Figure \ref{fig:attitude} (Right). Positive (cooperative) attitudes strongly correlate with higher emotional intensity, indicating that engaged patients are more willing to express emotions. In contrast, negative (resistant) attitudes correspond to lower and more dispersed emotional intensity. Modeling these attitude–emotion relationships enables more informed and adaptive response strategy selection.

\begin{figure*}[!t]
    \centering
    \includegraphics[height=0.25\textheight,width=\linewidth]{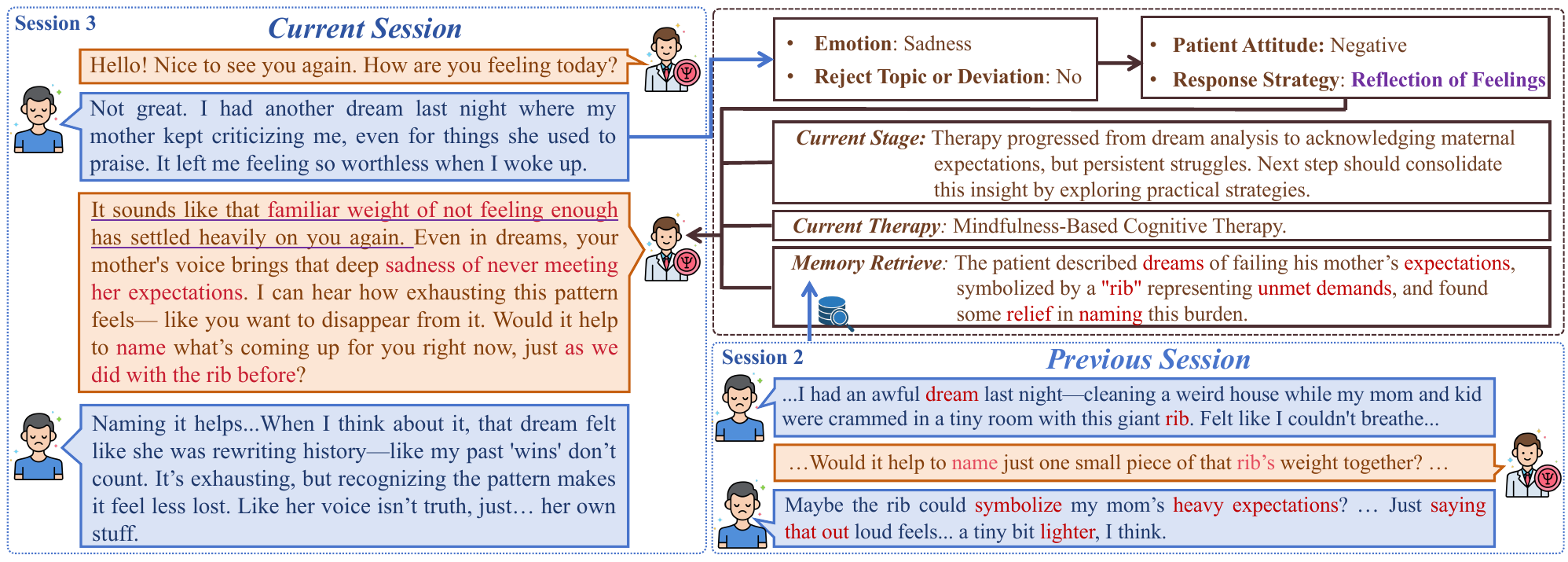}
    \vspace{-7mm}
    \caption{\small A case study of TheraMind's deliberative process. The agent synthesizes the patient's current state (emotion, attitude), the therapeutic stage, and a retrieved memory from a previous session (the ``rib''
metaphor) to generate a historically-informed, clinically-purposeful response.}
    \label{fig:casestudy}
    \vspace{-4mm}
\end{figure*}

\subsection{Case Study}
To qualitatively illustrate our agent's capabilities, Figure~\ref{fig:casestudy} presents a case study from Session 3. The patient describes a distressing dream about their mother, prompting TheraMind's deliberative response generation process.
First, the agent's intra-session loop perceives the patient's ``Sadness'' and ``Negative'' attitude, selecting a ``Reflection of Feelings'' strategy to validate their emotional state. Simultaneously, its cross-session memory module is triggered, retrieving a key insight from Session 2: a ``rib'' metaphor that the patient had previously linked to their mother's ``heavy expectations''.
The agent’s final response masterfully synthesizes these elements. It reflects the patient's immediate feelings (``familiar weight'', ``deep sadness'') while seamlessly weaving in the retrieved memory to create continuity (``...just as we did with the rib before?''). This single turn demonstrates how TheraMind avoids ``clinical amnesia''. By linking past breakthroughs to the present conversation, it transforms a simple reactive reply into a meaningful clinical intervention that reinforces the longitudinal therapeutic arc.

\section{Conclusion}
In this work, we introduced TheraMind, a strategic and adaptive agent that overcomes the critical limitations of prior work, clinical amnesia and strategic rigidity, through a novel dual-loop architecture. This framework decouples tactical, in-session dialogue management from strategic, cross-session therapeutic planning, enabling true longitudinal counseling. Extensive experiments and human evaluations confirmed its state-of-the-art performance, demonstrating that our architecture enables superior long-term coherence and flexibility.

%% file: WWW/appendix.tex
\clearpage
\section{Human Evaluation of Pairwise Response Selection}
\begin{figure}[H]
    \centering
    \includegraphics[width=\linewidth]{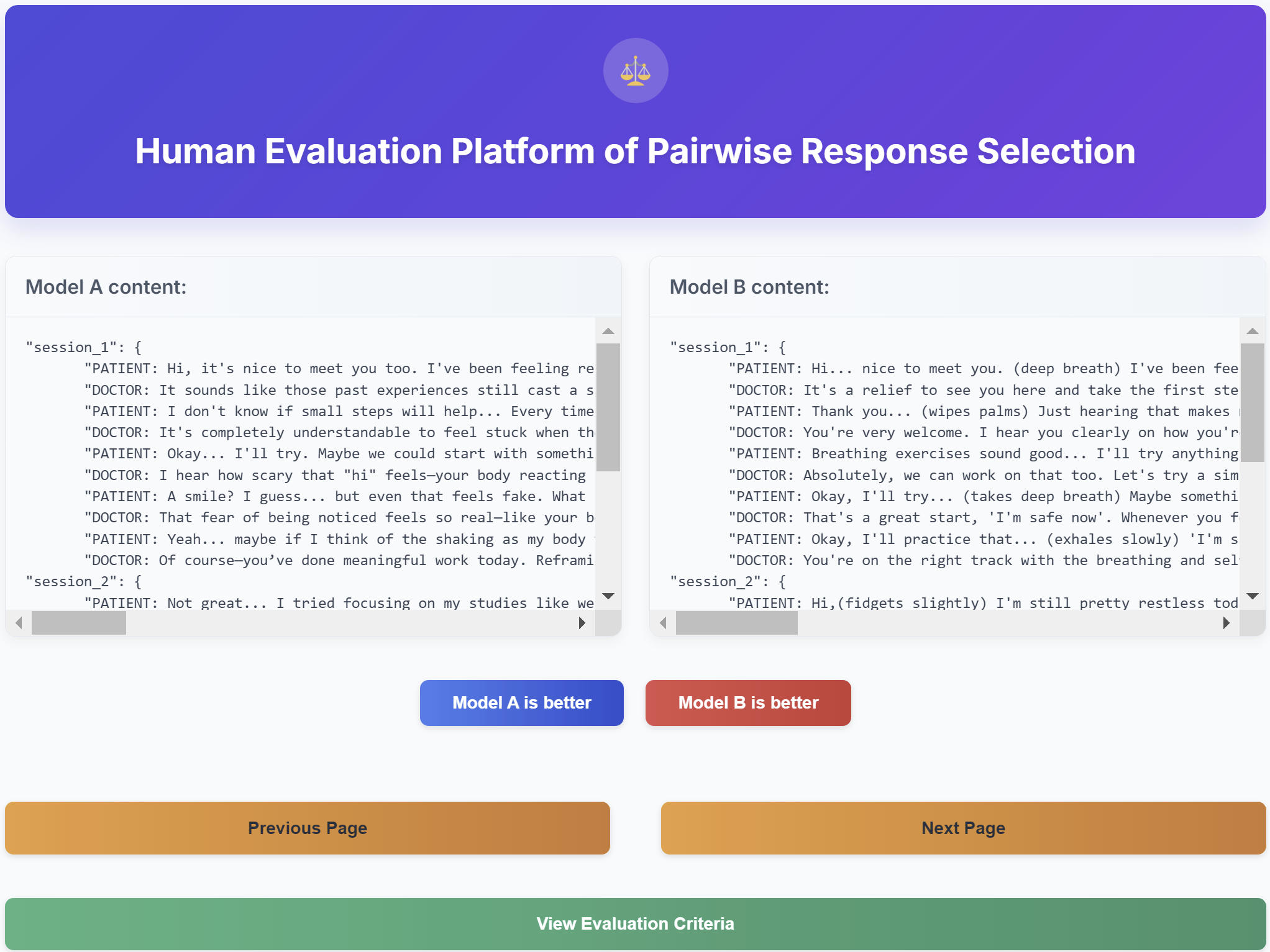}
    \caption{\small Human Evaluation Platform of Pairwise Response Performance Comparison.}
    \label{fig:humanpairwise}
\end{figure}

To ensure the universality of human evaluation, we randomly selected two cases from each of the ten categories as samples. Based on the predefined evaluation criteria, well-trained annotators respectively compared TheraMind with PsyDTLLM, DeepSeek-V3, and Interactive Agents in three pairwise scoring groups using our self-designed scoring platform, as presented in Figure~\ref{fig:humanpairwise}. By counting the winning situation of TheraMind with better response quality in the twenty samples (including 120 sessions in total), we confirm that our TheraMind has demonstrated better counseling effects in various psychological counseling scenarios. In addition, to verify the reliability of our human evaluation in model comparison evaluation, we used Cohen's~$\kappa$ coefficient to quantify the consistency of the evaluation results of multiple human evaluators. The final Cohen's $\kappa$ obtained was 0.697 (0.6--0.8), indicating that our human evaluation has high objectivity and impartiality, and the results of manual evaluation are generally reliable. 

\section{Human Agreement Analysis}
\begin{figure}[H]
    \centering
    \includegraphics[width=\linewidth]{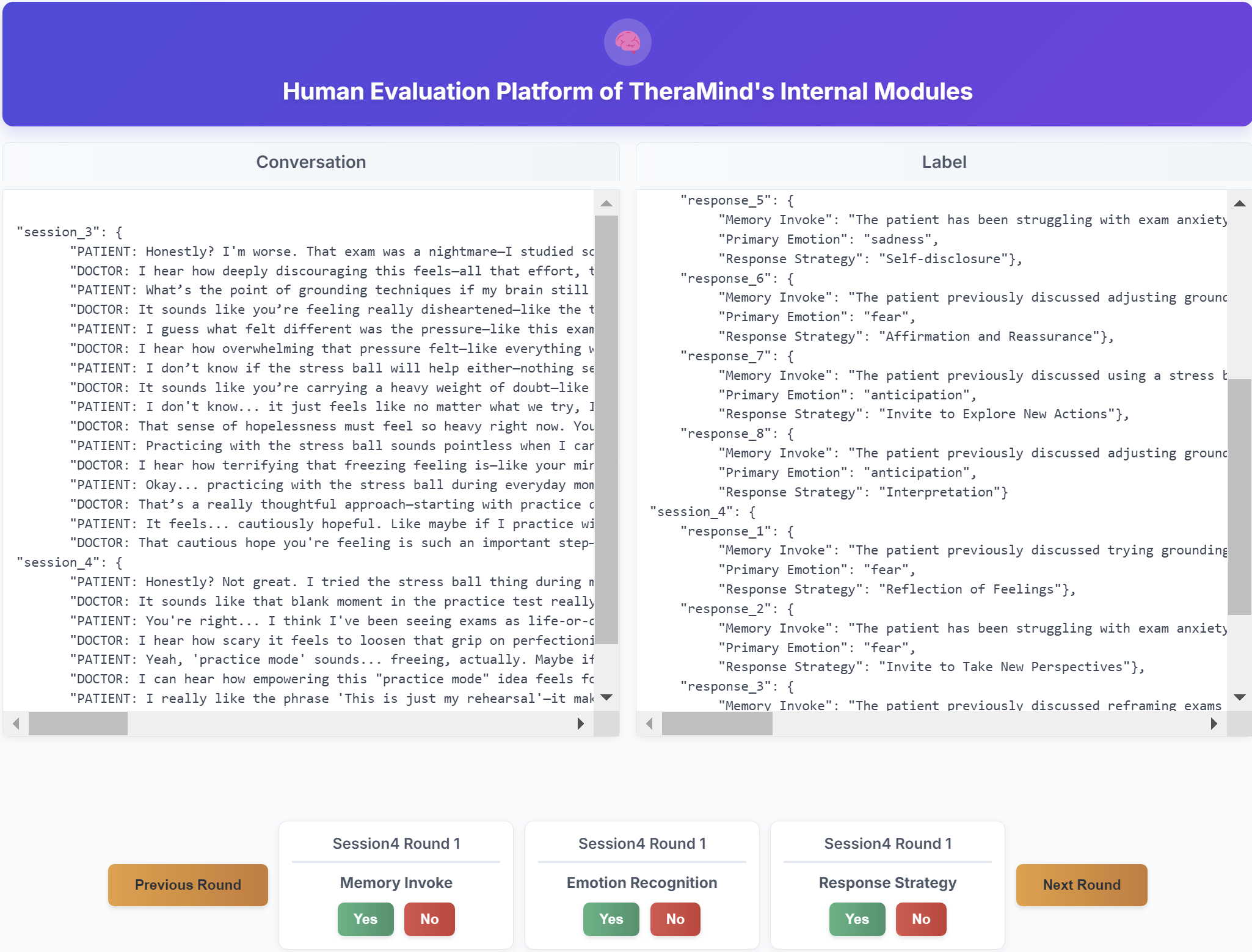}
    \caption{\small Human Evaluation Platform of TheraMind’s Internal Decision-making Modules.}
    \label{fig:humanmodules}
\end{figure}

To verify whether the three key decision-making modules of TheraMind, namely \textit{Memory Invoke}, \textit{Emotion Recognition}, and \textit{Response Strategy}, can effectively enhance counseling performance and accurately perform their intended judgments, two trained annotators with expertise in psychology evaluated the accuracy of these functions using our self-designed scoring platform, as shown in Figure~\ref{fig:humanmodules}. One case was randomly selected from each of the ten categories, and one session from each case was used for evaluation, covering a total of 68 dialogue rounds. To ensure a comprehensive assessment, the annotators were provided with both the target session and the previous sessions for contextual reference.

As shown in Figure~\ref{fig:agreement} (Left), TheraMind's modules demonstrate high fidelity with human clinical reasoning. The evaluation results show that the Emotion Recognition module reached an agreement rate of 95.59\%, and the Memory Invoke module achieved 94.12\%, demonstrating TheraMind’s strong perceptual and contextual reasoning abilities. The Response Strategy module also obtained an agreement rate of 85.29\%, indicating its capacity to make clinically coherent and contextually appropriate decisions. Inter-rater reliability, measured using Cohen’s $\kappa$ was calculated as 0.67, which falls within the range typically interpreted as substantial agreement (0.60--0.80). These findings confirm the reliability and consistency of TheraMind's core decision-making mechanisms in producing high-quality therapeutic responses.

\section{Ethical Considerations}
\begin{itemize}[left=0pt]
    \item \textbf{Data Privacy.} We used the CPsyCounR dataset of anonymized and professionally rewritten Chinese counseling reports strictly for academic research under ethical guidelines, with safeguards to prevent re-identification or misuse.
    \item \textbf{Annotator Salary.} All assessments were conducted by trained annotators with expertise in psychology, who were compensated fairly with payment set above local minimum wage and aligned with standard academic research practices.
    \item \textbf{Evaluation Bias.} Multi-session evaluation is challenging due to long dialogue histories. LLM-based methods offer scalability but may introduce biases (e.g., DeepSeek). We mitigated this with limited human expert assessments.
\end{itemize}

\section{Expert Evaluation}
To better demonstrate the advantages of our TheraMind, we randomly conducted a comparative analysis using the same therapeutic session from an identical case, comparing TheraMind with
PsyDTLLM which is a representative of our baseline models (see
Figure~\ref{fig:comparisontwo}). From the case, PsyDTLLM tends to overlook subtle emotional signals from patients, prematurely focuses on therapeutic
techniques without adequate emotional exploration, and sometimes
fails to effectively utilize historical dialogue information. These limitations lead to insufficient depth in emotional processing and poor
continuity of treatment due to memory consistency biases. In contrast, TheraMind demonstrates superior performance in multiple
aspects. Our framework excels in finely controlling therapeutic pace
and maintaining appropriate session boundaries. Crucially, TheraMind employs a patient-centered approach that prioritizes emotional understanding as the entry point for intervention. Building
upon empathic responses, it gradually guides patients to uncover
the root causes of their negative emotions, thereby enabling more
fundamental emotional resolution. The adaptive response strategies
employed by TheraMind allow for more natural pace adjustments
and better stage management of the therapeutic process.

\begin{figure*}[!b]
\begin{tcolorbox}[
    colback=myblue!5!white,
    colframe=myblue!75!black,
    arc=1mm, 
    auto outer arc,
    title={Counseling Performance Comparison},
    width=\textwidth
]
\footnotesize
\noindent 
\begin{tabular}{@{}p{5em}@{}p{\dimexpr\textwidth-5em-1\tabcolsep\relax}@{}}
\textbf{[TheraMind]}\\
PATIENT: &Hi... honestly, I'm feeling kind of mixed today. The anger about the bullying has been bubbling up, but it scares me—like if I let it out, I’ll lose control. I’ve been thinking a lot about how my grandparents raised me too... it feels tangled up with all this.\\
DOCTOR: &I hear how tangled this feels—anger that wants to surface but also terrifies you, tied to old wounds from bullying and your grandparents. It takes courage to sit with that. Let’s explore what control means to you. Could the anger be trying to protect you, like your body’s reactions did? Breathe with me for a moment—where do you feel it most right now?\\
PATIENT: &I feel it mostly in my chest—like a pressure. But talking about it makes me scared I’ll say something I regret. My grandparents always said anger was weak... I don’t know if it’s protecting me or just making things worse.\\
DOCTOR: &That pressure in your chest sounds like anger trying to be heard—not weakness, but a part of you that cares deeply. It’s okay to feel scared of regrets; let’s honor that caution while gently making space for this emotion. Would you be willing to notice the chest pressure together, just observing without judgment for a moment? Breathe with me.\\
PATIENT: &Okay, I'll try breathing with you... It's just hard when the pressure feels so heavy. But maybe if I think of it as my anger caring, not attacking... that helps a little.\\
DOCTOR: &I see. I really understand how heavy that pressure feels—it’s brave to sit with it. Let’s stay with this idea of anger caring for you. As we breathe, notice if the weight shifts even slightly. What’s one small thing your anger might be trying to protect you from right now? We’ll go at your pace.\\
PATIENT: &Maybe... it's trying to protect me from feeling that same helplessness I felt when the bullies laughed. Like if I stay angry, I won't get hurt again. That feels weird to say—but also kinda true?\\
DOCTOR: &That makes so much sense—your anger stepping in like a shield against that old helplessness. It’s not weird at all. Let’s sit with that truth for a moment. Can you notice how the anger feels now, knowing it’s trying to care for you? We’ll keep exploring this gently, just as it comes.\\
PATIENT: &It still feels dangerous... like if I let the anger protect me, I’ll turn into one of them—the people who hurt me. My grandparents’ voice in my head keeps saying ‘good boys don’t get angry.’ Maybe we should stop here... this is too much.\\
DOCTOR: &I can empathize with your feelings—like standing at the edge of something scary. Your anger isn’t the bullies; it’s just a part of you that learned to shout when you couldn’t speak. And that voice about ‘good boys’? Let’s breathe together for a moment, just noticing it without fighting. You’re safe here. Would you like to pause or keep exploring gently?\\
PATIENT: &Yeah... breathing helps a little. But I still hate that I even have to deal with this anger—it's not fair. I'll try to hold onto that idea about protection though. I think I'd like to pause here today. I need time to process.\\
DOCTOR: &Of course—it’s completely understandable to feel frustrated about carrying this anger. You’re honoring your needs by taking time to process, and that’s important. Let’s pause here. Remember, this anger has been trying to shield you, not define you. I’ll be here when you’re ready to continue. Take gentle care until then.\\
\textbf{[PsyDTLLM]}\\
 PATIENT: &Not great... (shifts uncomfortably) I've been thinking a lot about the kids who bullied me. It makes me so angry but... I don't know what to do with that anger. Part of me just wants to forget it ever happened.\\
DOCTOR: &Your experience was terrible and it's understandable that it still affects you. It's not always easy to forgive, especially when it comes to such a difficult time in your life. It's okay to feel those emotions, and it's equally okay to want to move past them. We can work on strategies to manage those feelings if you're ready to take that step.\\
PATIENT: &Yeah... (fidgets) I think I need to talk about my childhood with my grandparents. They tried their best, but I always felt...lonely. Maybe that's why the bullying hurt so much. I want to understand how these pieces connect.\\
DOCTOR: &It's clear that your grandparents played a significant role in your formative years. Their influence, combined with the bullying, leaves a lasting impact. Would you like to explore how your early experiences have shaped your current feelings and behaviors?\\
PATIENT: &Yeah... (takes shaky breath) My grandparents were loving but strict - they didn't know how to help with the bullying. I think that's why I learned to swallow my anger. It's scary to admit how much it still hurts, but I want to understand this better.\\
DOCTOR: &Acknowledging and understanding the pain from your past is a brave step. Swallowing anger can be a coping mechanism but also a source of hidden suffering. Let's explore how we can bring that anger to the surface and find a way to heal it together.\\
PATIENT: &Okay... (clenches fists slightly) I think I'm ready to try expressing that anger. Maybe starting with writing letters I'll never send? I want to understand how being raised to 'stay quiet' connects to my bottled-up feelings now.\\
DOCTOR: &Writing those letters can be a powerful way to express your suppressed emotions. It's a means of giving voice to your feelings without the risk of confrontation. This can be a therapeutic process that allows you to understand the depth of your pain and anger.\\
PATIENT: &(shakes head) I don't know... writing letters feels scary. What if I get too angry? My grandparents always said 'don't make waves' and now I can't even tell if my anger is okay. Maybe we should stop here for today... (looks at clock) I need time to think about this.\\
DOCTOR: &Of course, take all the time you need. It's important to feel comfortable with the steps we take together. We'll resume when you're ready.\\
PATIENT: &(sighs heavily) I just... I don't think I can do this today. The idea of digging up all that old pain makes my chest feel tight. Maybe next time we can start slower? I need to go now.\\
DOCTOR: &That's okay, we can go at your pace. Please take care of yourself and we'll continue our work when you're ready.\\
PATIENT: &(shakes head) I don't think I can handle this today... maybe we should just stop. The more we talk about my grandparents and the bullying, the more my stomach knots up. I need to go.\\
DOCTOR: &Understood, we can pause here. It's important to respect how you feel in the moment. We'll continue when you're ready and feel more comfortable. Take care of yourself until then.\\
\end{tabular}
\end{tcolorbox}
\caption{Comparision of TheraMind's and PsyDTLLM's Counseling Performances.}
\label{fig:comparisontwo}
\end{figure*}

%% file: WWW/sample-sigconf-authordraft.bbl

\begin{thebibliography}{59}


\ifx \showCODEN    \undefined \def \showCODEN     #1{\unskip}     \fi
\ifx \showISBNx    \undefined \def \showISBNx     #1{\unskip}     \fi
\ifx \showISBNxiii \undefined \def \showISBNxiii  #1{\unskip}     \fi
\ifx \showISSN     \undefined \def \showISSN      #1{\unskip}     \fi
\ifx \showLCCN     \undefined \def \showLCCN      #1{\unskip}     \fi
\ifx \shownote     \undefined \def \shownote      #1{#1}          \fi
\ifx \showarticletitle \undefined \def \showarticletitle #1{#1}   \fi
\ifx \showURL      \undefined \def \showURL       {\relax}        \fi
\providecommand\bibfield[2]{#2}
\providecommand\bibinfo[2]{#2}
\providecommand\natexlab[1]{#1}
\providecommand\showeprint[2][]{arXiv:#2}

\bibitem[Anil et~al\mbox{.}(2023)]%
        {Gemini}
\bibfield{author}{\bibinfo{person}{Rohan Anil}, \bibinfo{person}{Sebastian Borgeaud}, \bibinfo{person}{Yonghui Wu}, \bibinfo{person}{Jean{-}Baptiste Alayrac}, \bibinfo{person}{Jiahui Yu}, \bibinfo{person}{Radu Soricut}, \bibinfo{person}{Johan Schalkwyk}, \bibinfo{person}{Andrew~M. Dai}, \bibinfo{person}{Anja Hauth}, \bibinfo{person}{Katie Millican}, \bibinfo{person}{David Silver}, \bibinfo{person}{Slav Petrov}, \bibinfo{person}{Melvin Johnson}, \bibinfo{person}{Ioannis Antonoglou}, \bibinfo{person}{Julian Schrittwieser}, \bibinfo{person}{Amelia Glaese}, \bibinfo{person}{Jilin Chen}, \bibinfo{person}{Emily Pitler}, \bibinfo{person}{Timothy~P. Lillicrap}, \bibinfo{person}{Angeliki Lazaridou}, \bibinfo{person}{Orhan Firat}, \bibinfo{person}{James Molloy}, \bibinfo{person}{Michael Isard}, \bibinfo{person}{Paul~Ronald Barham}, \bibinfo{person}{Tom Hennigan}, \bibinfo{person}{Benjamin Lee}, \bibinfo{person}{Fabio Viola}, \bibinfo{person}{Malcolm Reynolds}, \bibinfo{person}{Yuanzhong Xu}, \bibinfo{person}{Ryan
  Doherty}, \bibinfo{person}{Eli Collins}, \bibinfo{person}{Clemens Meyer}, \bibinfo{person}{Eliza Rutherford}, \bibinfo{person}{Erica Moreira}, \bibinfo{person}{Kareem Ayoub}, \bibinfo{person}{Megha Goel}, \bibinfo{person}{George Tucker}, \bibinfo{person}{Enrique Piqueras}, \bibinfo{person}{Maxim Krikun}, \bibinfo{person}{Iain Barr}, \bibinfo{person}{Nikolay Savinov}, \bibinfo{person}{Ivo Danihelka}, \bibinfo{person}{Becca Roelofs}, \bibinfo{person}{Ana{\"{\i}}s White}, \bibinfo{person}{Anders Andreassen}, \bibinfo{person}{Tamara von Glehn}, \bibinfo{person}{Lakshman Yagati}, \bibinfo{person}{Mehran Kazemi}, \bibinfo{person}{Lucas Gonzalez}, \bibinfo{person}{Misha Khalman}, \bibinfo{person}{Jakub Sygnowski}, {and} \bibinfo{person}{et al.}} \bibinfo{year}{2023}\natexlab{}.
\newblock \bibinfo{title}{Gemini: {A} Family of Highly Capable Multimodal Models}.
\newblock
\showeprint[arXiv]{2312.11805}


\bibitem[Beck(2011)]%
        {CBT}
\bibfield{author}{\bibinfo{person}{Judith~S Beck}.} \bibinfo{year}{2011}\natexlab{}.
\newblock \showarticletitle{Cognitive-behavioral therapy}.
\newblock \bibinfo{journal}{\emph{Clinical textbook of addictive disorders}}  \bibinfo{volume}{491} (\bibinfo{year}{2011}), \bibinfo{pages}{474--501}.
\newblock


\bibitem[Bi et~al\mbox{.}(2025)]%
        {MAGI}
\bibfield{author}{\bibinfo{person}{Guanqun Bi}, \bibinfo{person}{Zhuang Chen}, \bibinfo{person}{Zhoufu Liu}, \bibinfo{person}{Hongkai Wang}, \bibinfo{person}{Xiyao Xiao}, \bibinfo{person}{Yuqiang Xie}, \bibinfo{person}{Wen Zhang}, \bibinfo{person}{Yongkang Huang}, \bibinfo{person}{Yuxuan Chen}, \bibinfo{person}{Libiao Peng}, {and} \bibinfo{person}{Minlie Huang}.} \bibinfo{year}{2025}\natexlab{}.
\newblock \showarticletitle{{MAGI:} Multi-Agent Guided Interview for Psychiatric Assessment}. In \bibinfo{booktitle}{\emph{Findings of the Association for Computational Linguistics, {ACL} 2025, Vienna, Austria, July 27 - August 1, 2025}}. \bibinfo{address}{Vienna, Austria}, \bibinfo{pages}{24898--24921}.
\newblock


\bibitem[Bo et~al\mbox{.}(2024)]%
        {Multi-Agent1}
\bibfield{author}{\bibinfo{person}{Xiaohe Bo}, \bibinfo{person}{Zeyu Zhang}, \bibinfo{person}{Quanyu Dai}, \bibinfo{person}{Xueyang Feng}, \bibinfo{person}{Lei Wang}, \bibinfo{person}{Rui Li}, \bibinfo{person}{Xu Chen}, {and} \bibinfo{person}{Ji{-}Rong Wen}.} \bibinfo{year}{2024}\natexlab{}.
\newblock \showarticletitle{Reflective Multi-Agent Collaboration based on Large Language Models}. In \bibinfo{booktitle}{\emph{Advances in Neural Information Processing Systems 38: Annual Conference on Neural Information Processing Systems 2024, NeurIPS 2024, Vancouver, BC, Canada, December 10 - 15, 2024}}. \bibinfo{publisher}{NeurIPS}, \bibinfo{address}{Vancouver, BC, Canada}.
\newblock


\bibitem[Cai et~al\mbox{.}(2024)]%
        {cai2024internlm2}
\bibfield{author}{\bibinfo{person}{Zheng Cai}, \bibinfo{person}{Maosong Cao}, \bibinfo{person}{Haojiong Chen}, \bibinfo{person}{Kai Chen}, \bibinfo{person}{Keyu Chen}, \bibinfo{person}{Xin Chen}, \bibinfo{person}{Xun Chen}, \bibinfo{person}{Zehui Chen}, \bibinfo{person}{Zhi Chen}, \bibinfo{person}{Pei Chu}, {et~al\mbox{.}}} \bibinfo{year}{2024}\natexlab{}.
\newblock \bibinfo{title}{Internlm2 technical report}.
\newblock


\bibitem[Chen et~al\mbox{.}(2025)]%
        {MIND}
\bibfield{author}{\bibinfo{person}{Yujia Chen}, \bibinfo{person}{Changsong Li}, \bibinfo{person}{Yiming Wang}, \bibinfo{person}{Qingqing Xiao}, \bibinfo{person}{Nan Zhang}, \bibinfo{person}{Zifan Kong}, \bibinfo{person}{Peng Wang}, {and} \bibinfo{person}{Binyu Yan}.} \bibinfo{year}{2025}\natexlab{}.
\newblock \bibinfo{title}{{MIND:} Towards Immersive Psychological Healing with Multi-agent Inner Dialogue}.
\newblock
\showeprint[arXiv]{2502.19860}


\bibitem[Cohen(1960)]%
        {cohens}
\bibfield{author}{\bibinfo{person}{Jacob Cohen}.} \bibinfo{year}{1960}\natexlab{}.
\newblock \showarticletitle{A coefficient of agreement for nominal scales}.
\newblock \bibinfo{journal}{\emph{Educational and psychological measurement}} \bibinfo{volume}{20}, \bibinfo{number}{1} (\bibinfo{year}{1960}), \bibinfo{pages}{37--46}.
\newblock


\bibitem[Comai et~al\mbox{.}(2025)]%
        {comai2025moving}
\bibfield{author}{\bibinfo{person}{Stefano Comai}, \bibinfo{person}{Mirko Manchia}, \bibinfo{person}{Marta Bosia}, \bibinfo{person}{Alessandro Miola}, \bibinfo{person}{Sara Poletti}, \bibinfo{person}{Francesco Benedetti}, \bibinfo{person}{Sofia Nasini}, \bibinfo{person}{Raffaele Ferri}, \bibinfo{person}{Dan Rujescu}, \bibinfo{person}{Marion Leboyer}, {et~al\mbox{.}}} \bibinfo{year}{2025}\natexlab{}.
\newblock \showarticletitle{Moving toward precision and personalized treatment strategies in psychiatry}.
\newblock \bibinfo{journal}{\emph{International Journal of Neuropsychopharmacology}} \bibinfo{volume}{28}, \bibinfo{number}{5} (\bibinfo{year}{2025}), \bibinfo{pages}{pyaf025}.
\newblock


\bibitem[Doran(2016)]%
        {working}
\bibfield{author}{\bibinfo{person}{Jennifer~M Doran}.} \bibinfo{year}{2016}\natexlab{}.
\newblock \showarticletitle{The working alliance: Where have we been, where are we going?}
\newblock \bibinfo{journal}{\emph{Psychotherapy Research}} \bibinfo{volume}{26}, \bibinfo{number}{2} (\bibinfo{year}{2016}), \bibinfo{pages}{146--163}.
\newblock


\bibitem[Ge et~al\mbox{.}(2025)]%
        {Survey2}
\bibfield{author}{\bibinfo{person}{Zhuohan Ge}, \bibinfo{person}{Nicole Hu}, \bibinfo{person}{Darian Li}, \bibinfo{person}{Yubo Wang}, \bibinfo{person}{Shihao Qi}, \bibinfo{person}{Yuming Xu}, \bibinfo{person}{Han Shi}, {and} \bibinfo{person}{Jason Zhang}.} \bibinfo{year}{2025}\natexlab{}.
\newblock \bibinfo{title}{A Survey of Large Language Models in Mental Health Disorder Detection on Social Media}.
\newblock
\showeprint[arXiv]{2504.02800}


\bibitem[Ghosh et~al\mbox{.}(2024)]%
        {Closer}
\bibfield{author}{\bibinfo{person}{Sreyan Ghosh}, \bibinfo{person}{Chandra Kiran~Reddy Evuru}, \bibinfo{person}{Sonal Kumar}, \bibinfo{person}{Ramaneswaran S.}, \bibinfo{person}{Deepali Aneja}, \bibinfo{person}{Zeyu Jin}, \bibinfo{person}{Ramani Duraiswami}, {and} \bibinfo{person}{Dinesh Manocha}.} \bibinfo{year}{2024}\natexlab{}.
\newblock \showarticletitle{A Closer Look at the Limitations of Instruction Tuning}. In \bibinfo{booktitle}{\emph{Forty-first International Conference on Machine Learning, {ICML} 2024, Vienna, Austria, July 21-27, 2024}}. \bibinfo{publisher}{OpenReview.net}, \bibinfo{address}{Vienna, Austria}.
\newblock


\bibitem[GLM et~al\mbox{.}(2024)]%
        {glm2024chatglm}
\bibfield{author}{\bibinfo{person}{Team GLM}, \bibinfo{person}{Aohan Zeng}, \bibinfo{person}{Bin Xu}, \bibinfo{person}{Bowen Wang}, \bibinfo{person}{Chenhui Zhang}, \bibinfo{person}{Da Yin}, \bibinfo{person}{Dan Zhang}, \bibinfo{person}{Diego Rojas}, \bibinfo{person}{Guanyu Feng}, \bibinfo{person}{Hanlin Zhao}, {et~al\mbox{.}}} \bibinfo{year}{2024}\natexlab{}.
\newblock \bibinfo{title}{Chatglm: A family of large language models from glm-130b to glm-4 all tools}.
\newblock


\bibitem[Guo et~al\mbox{.}(2025)]%
        {guo2025seed1}
\bibfield{author}{\bibinfo{person}{Dong Guo}, \bibinfo{person}{Faming Wu}, \bibinfo{person}{Feida Zhu}, \bibinfo{person}{Fuxing Leng}, \bibinfo{person}{Guang Shi}, \bibinfo{person}{Haobin Chen}, \bibinfo{person}{Haoqi Fan}, \bibinfo{person}{Jian Wang}, \bibinfo{person}{Jianyu Jiang}, \bibinfo{person}{Jiawei Wang}, {et~al\mbox{.}}} \bibinfo{year}{2025}\natexlab{}.
\newblock \bibinfo{title}{Seed1. 5-vl technical report}.
\newblock


\bibitem[He et~al\mbox{.}(2021)]%
        {Speaker}
\bibfield{author}{\bibinfo{person}{Zihao He}, \bibinfo{person}{Leili Tavabi}, \bibinfo{person}{Kristina Lerman}, {and} \bibinfo{person}{Mohammad Soleymani}.} \bibinfo{year}{2021}\natexlab{}.
\newblock \showarticletitle{Speaker Turn Modeling for Dialogue Act Classification}. In \bibinfo{booktitle}{\emph{Findings of the Association for Computational Linguistics: {EMNLP} 2021, Virtual Event / Punta Cana, Dominican Republic, 16-20 November, 2021}}. \bibinfo{publisher}{Association for Computational Linguistics}, \bibinfo{address}{Virtual Event / Punta Cana, Dominican Republic}, \bibinfo{pages}{2150--2157}.
\newblock


\bibitem[Hill(1999)]%
        {hill1999helping}
\bibfield{author}{\bibinfo{person}{Clara~E Hill}.} \bibinfo{year}{1999}\natexlab{}.
\newblock \showarticletitle{Helping skills: Facilitating exploration, insight, and action}.
\newblock \bibinfo{journal}{\emph{American Psychological Association}} (\bibinfo{year}{1999}).
\newblock


\bibitem[Hu et~al\mbox{.}(2025a)]%
        {Beyond}
\bibfield{author}{\bibinfo{person}{He Hu}, \bibinfo{person}{Yucheng Zhou}, \bibinfo{person}{Juzheng Si}, \bibinfo{person}{Qianning Wang}, \bibinfo{person}{Hengheng Zhang}, \bibinfo{person}{Fuji Ren}, \bibinfo{person}{Fei Ma}, {and} \bibinfo{person}{Laizhong Cui}.} \bibinfo{year}{2025}\natexlab{a}.
\newblock \bibinfo{title}{Beyond Empathy: Integrating Diagnostic and Therapeutic Reasoning with Large Language Models for Mental Health Counseling}.
\newblock
\showeprint[arXiv]{2505.15715}


\bibitem[Hu et~al\mbox{.}(2025b)]%
        {pattern}
\bibfield{author}{\bibinfo{person}{He Hu}, \bibinfo{person}{Yucheng Zhou}, \bibinfo{person}{Qianning Wang}, \bibinfo{person}{Yingjian Zou}, \bibinfo{person}{Chiyuan Ma}, \bibinfo{person}{Juzheng Si}, \bibinfo{person}{Jianzhuang Liu}, \bibinfo{person}{Zitong Yu}, \bibinfo{person}{Laizhong Cui}, {and} \bibinfo{person}{Fei Ma}.} \bibinfo{year}{2025}\natexlab{b}.
\newblock \showarticletitle{From Pattern Recognizers to Personalized Companions: A Survey of Large Language Models in Mental Health}.
\newblock  (\bibinfo{year}{2025}).
\newblock


\bibitem[Hu et~al\mbox{.}(2025c)]%
        {EmoBench}
\bibfield{author}{\bibinfo{person}{He Hu}, \bibinfo{person}{Yucheng Zhou}, \bibinfo{person}{Lianzhong You}, \bibinfo{person}{Hongbo Xu}, \bibinfo{person}{Qianning Wang}, \bibinfo{person}{Zheng Lian}, \bibinfo{person}{Fei~Richard Yu}, \bibinfo{person}{Fei Ma}, {and} \bibinfo{person}{Laizhong Cui}.} \bibinfo{year}{2025}\natexlab{c}.
\newblock \showarticletitle{EmoBench-M: Benchmarking Emotional Intelligence for Multimodal Large Language Models}.
\newblock \bibinfo{journal}{\emph{CoRR}}  \bibinfo{volume}{abs/2502.04424} (\bibinfo{year}{2025}).
\newblock
\showeprint[arXiv]{2502.04424}


\bibitem[Hua et~al\mbox{.}(2025)]%
        {scoping}
\bibfield{author}{\bibinfo{person}{Yining Hua}, \bibinfo{person}{Hongbin Na}, \bibinfo{person}{Zehan Li}, \bibinfo{person}{Fenglin Liu}, \bibinfo{person}{Xiao Fang}, \bibinfo{person}{David~A. Clifton}, {and} \bibinfo{person}{John~B. Torous}.} \bibinfo{year}{2025}\natexlab{}.
\newblock \showarticletitle{A scoping review of large language models for generative tasks in mental health care}.
\newblock \bibinfo{journal}{\emph{npj Digit. Medicine}} \bibinfo{volume}{8}, \bibinfo{number}{1} (\bibinfo{year}{2025}).
\newblock


\bibitem[Ji et~al\mbox{.}(2025)]%
        {Transformin_Mental}
\bibfield{author}{\bibinfo{person}{Sijie Ji}, \bibinfo{person}{Xinzhe Zheng}, \bibinfo{person}{Wei Gao}, {and} \bibinfo{person}{Mani Srivastava}.} \bibinfo{year}{2025}\natexlab{}.
\newblock \showarticletitle{Transforming Mental Health Care with Autonomous {LLM} Agents at the Edge}. In \bibinfo{booktitle}{\emph{Proceedings of the 23rd {ACM} Conference on Embedded Networked Sensor Systems, SenSys 2025, {UC} Irvine Student Center, Irvine, CA, USA, May 6-9, 2025}}. \bibinfo{publisher}{{ACM}}, \bibinfo{address}{Irvine, CA, USA}, \bibinfo{pages}{692--693}.
\newblock


\bibitem[Kwon et~al\mbox{.}(2023)]%
        {Reward}
\bibfield{author}{\bibinfo{person}{Minae Kwon}, \bibinfo{person}{Sang~Michael Xie}, \bibinfo{person}{Kalesha Bullard}, {and} \bibinfo{person}{Dorsa Sadigh}.} \bibinfo{year}{2023}\natexlab{}.
\newblock \showarticletitle{Reward Design with Language Models}. In \bibinfo{booktitle}{\emph{The Eleventh International Conference on Learning Representations, {ICLR} 2023, Kigali, Rwanda, May 1-5, 2023}}. \bibinfo{publisher}{OpenReview.net}, \bibinfo{address}{Kigali, Rwanda}.
\newblock


\bibitem[Lee et~al\mbox{.}(2024)]%
        {catus}
\bibfield{author}{\bibinfo{person}{Suyeon Lee}, \bibinfo{person}{Sunghwan Kim}, \bibinfo{person}{Minju Kim}, \bibinfo{person}{Dongjin Kang}, \bibinfo{person}{Dongil Yang}, \bibinfo{person}{Harim Kim}, \bibinfo{person}{Minseok Kang}, \bibinfo{person}{Dayi Jung}, \bibinfo{person}{Min~Hee Kim}, \bibinfo{person}{Seungbeen Lee}, \bibinfo{person}{Kyoung{-}Mee Chung}, \bibinfo{person}{Youngjae Yu}, \bibinfo{person}{Dongha Lee}, {and} \bibinfo{person}{Jinyoung Yeo}.} \bibinfo{year}{2024}\natexlab{}.
\newblock \showarticletitle{Cactus: Towards Psychological Counseling Conversations using Cognitive Behavioral Theory}. In \bibinfo{booktitle}{\emph{Findings of the Association for Computational Linguistics: {EMNLP} 2024, Miami, Florida, USA, November 12-16, 2024}}. \bibinfo{publisher}{Association for Computational Linguistics}, \bibinfo{address}{Miami, Florida, USA}, \bibinfo{pages}{14245--14274}.
\newblock


\bibitem[Liu et~al\mbox{.}(2024a)]%
        {liu2024deepseek}
\bibfield{author}{\bibinfo{person}{Aixin Liu}, \bibinfo{person}{Bei Feng}, \bibinfo{person}{Bing Xue}, \bibinfo{person}{Bingxuan Wang}, \bibinfo{person}{Bochao Wu}, \bibinfo{person}{Chengda Lu}, \bibinfo{person}{Chenggang Zhao}, \bibinfo{person}{Chengqi Deng}, \bibinfo{person}{Chenyu Zhang}, \bibinfo{person}{Chong Ruan}, {et~al\mbox{.}}} \bibinfo{year}{2024}\natexlab{a}.
\newblock \bibinfo{title}{Deepseek-v3 technical report}.
\newblock


\bibitem[Liu et~al\mbox{.}(2023)]%
        {chatcounselor}
\bibfield{author}{\bibinfo{person}{June~M Liu}, \bibinfo{person}{Donghao Li}, \bibinfo{person}{He Cao}, \bibinfo{person}{Tianhe Ren}, \bibinfo{person}{Zeyi Liao}, {and} \bibinfo{person}{Jiamin Wu}.} \bibinfo{year}{2023}\natexlab{}.
\newblock \bibinfo{title}{Chatcounselor: A large language models for mental health support}.
\newblock


\bibitem[Liu et~al\mbox{.}(2024b)]%
        {AgentBench}
\bibfield{author}{\bibinfo{person}{Xiao Liu}, \bibinfo{person}{Hao Yu}, \bibinfo{person}{Hanchen Zhang}, \bibinfo{person}{Yifan Xu}, \bibinfo{person}{Xuanyu Lei}, \bibinfo{person}{Hanyu Lai}, \bibinfo{person}{Yu Gu}, \bibinfo{person}{Hangliang Ding}, \bibinfo{person}{Kaiwen Men}, \bibinfo{person}{Kejuan Yang}, \bibinfo{person}{Shudan Zhang}, \bibinfo{person}{Xiang Deng}, \bibinfo{person}{Aohan Zeng}, \bibinfo{person}{Zhengxiao Du}, \bibinfo{person}{Chenhui Zhang}, \bibinfo{person}{Sheng Shen}, \bibinfo{person}{Tianjun Zhang}, \bibinfo{person}{Yu Su}, \bibinfo{person}{Huan Sun}, \bibinfo{person}{Minlie Huang}, \bibinfo{person}{Yuxiao Dong}, {and} \bibinfo{person}{Jie Tang}.} \bibinfo{year}{2024}\natexlab{b}.
\newblock \showarticletitle{AgentBench: Evaluating LLMs as Agents}. In \bibinfo{booktitle}{\emph{The Twelfth International Conference on Learning Representations, {ICLR} 2024, Vienna, Austria, May 7-11, 2024}}. \bibinfo{publisher}{OpenReview.net}, \bibinfo{address}{Vienna, Austria}.
\newblock


\bibitem[Mao et~al\mbox{.}(2025)]%
        {EmoAgent}
\bibfield{author}{\bibinfo{person}{Qi Mao}, \bibinfo{person}{Haobo Hu}, \bibinfo{person}{Yujie He}, \bibinfo{person}{Difei Gao}, \bibinfo{person}{Haokun Chen}, {and} \bibinfo{person}{Libiao Jin}.} \bibinfo{year}{2025}\natexlab{}.
\newblock \bibinfo{title}{EmoAgent: Multi-Agent Collaboration of Plan, Edit, and Critic, for Affective Image Manipulation}.
\newblock
\showeprint[arXiv]{2503.11290}


\bibitem[Moggia et~al\mbox{.}(2024)]%
        {treatment}
\bibfield{author}{\bibinfo{person}{Danilo Moggia}, \bibinfo{person}{Wolfgang Lutz}, \bibinfo{person}{Eva-Lotta Brakemeier}, {and} \bibinfo{person}{Leonard Bickman}.} \bibinfo{year}{2024}\natexlab{}.
\newblock \showarticletitle{Treatment Personalization and Precision Mental Health Care: Where are we and where do we want to go?}
\newblock \bibinfo{journal}{\emph{Administration and Policy in Mental Health and Mental Health Services Research}} \bibinfo{volume}{51}, \bibinfo{number}{5} (\bibinfo{year}{2024}), \bibinfo{pages}{611--616}.
\newblock


\bibitem[Na et~al\mbox{.}(2025)]%
        {Survey1}
\bibfield{author}{\bibinfo{person}{Hongbin Na}, \bibinfo{person}{Yining Hua}, \bibinfo{person}{Zimu Wang}, \bibinfo{person}{Tao Shen}, \bibinfo{person}{Beibei Yu}, \bibinfo{person}{Lilin Wang}, \bibinfo{person}{Wei Wang}, \bibinfo{person}{John~B. Torous}, {and} \bibinfo{person}{Ling Chen}.} \bibinfo{year}{2025}\natexlab{}.
\newblock \showarticletitle{A Survey of Large Language Models in Psychotherapy: Current Landscape and Future Directions}. In \bibinfo{booktitle}{\emph{Findings of the Association for Computational Linguistics, {ACL} 2025, Vienna, Austria, July 27 - August 1, 2025}}. \bibinfo{publisher}{Association for Computational Linguistics}, \bibinfo{address}{Vienna, Austria}, \bibinfo{pages}{7362--7376}.
\newblock


\bibitem[OpenAI(2023)]%
        {GPT-4}
\bibfield{author}{\bibinfo{person}{OpenAI}.} \bibinfo{year}{2023}\natexlab{}.
\newblock \bibinfo{title}{{GPT-4} Technical Report}.
\newblock
\showeprint[arXiv]{2303.08774}


\bibitem[Poria et~al\mbox{.}(2019)]%
        {MELD}
\bibfield{author}{\bibinfo{person}{Soujanya Poria}, \bibinfo{person}{Devamanyu Hazarika}, \bibinfo{person}{Navonil Majumder}, \bibinfo{person}{Gautam Naik}, \bibinfo{person}{Erik Cambria}, {and} \bibinfo{person}{Rada Mihalcea}.} \bibinfo{year}{2019}\natexlab{}.
\newblock \showarticletitle{{MELD}: A Multimodal Multi-Party Dataset for Emotion Recognition in Conversations}. In \bibinfo{booktitle}{\emph{Proceedings of the 57th Annual Meeting of the Association for Computational Linguistics}}. \bibinfo{publisher}{Association for Computational Linguistics}, \bibinfo{address}{Florence, Italy}, \bibinfo{pages}{527--536}.
\newblock


\bibitem[Psy525(2007)]%
        {psy525}
\bibfield{author}{\bibinfo{person}{Psy525}.} \bibinfo{year}{2007}\natexlab{}.
\newblock \bibinfo{title}{Psychological Counseling and Mental Health Services}.
\newblock
\urldef\tempurl%
\url{https://www.psy525.cn}
\showURL{%
\tempurl}
\newblock
\shownote{Accessed: 2025-10-02}.


\bibitem[Qi et~al\mbox{.}(2025)]%
        {KokoroChat}
\bibfield{author}{\bibinfo{person}{Zhiyang Qi}, \bibinfo{person}{Takumasa Kaneko}, \bibinfo{person}{Keiko Takamizo}, \bibinfo{person}{Mariko Ukiyo}, {and} \bibinfo{person}{Michimasa Inaba}.} \bibinfo{year}{2025}\natexlab{}.
\newblock \showarticletitle{KokoroChat: {A} Japanese Psychological Counseling Dialogue Dataset Collected via Role-Playing by Trained Counselors}. In \bibinfo{booktitle}{\emph{Proceedings of the 63rd Annual Meeting of the Association for Computational Linguistics (Volume 1: Long Papers), {ACL} 2025, Vienna, Austria, July 27 - August 1, 2025}}. \bibinfo{publisher}{Association for Computational Linguistics}, \bibinfo{address}{Vienna, Austria}, \bibinfo{pages}{12424--12443}.
\newblock


\bibitem[Qiu et~al\mbox{.}(2024)]%
        {smile}
\bibfield{author}{\bibinfo{person}{Huachuan Qiu}, \bibinfo{person}{Hongliang He}, \bibinfo{person}{Shuai Zhang}, \bibinfo{person}{Anqi Li}, {and} \bibinfo{person}{Zhenzhong Lan}.} \bibinfo{year}{2024}\natexlab{}.
\newblock \showarticletitle{{SMILE:} Single-turn to Multi-turn Inclusive Language Expansion via ChatGPT for Mental Health Support}. In \bibinfo{booktitle}{\emph{Findings of the Association for Computational Linguistics: {EMNLP} 2024, Miami, Florida, USA, November 12-16, 2024}}. \bibinfo{publisher}{Association for Computational Linguistics}, \bibinfo{address}{Miami, Florida, USA}, \bibinfo{pages}{615--636}.
\newblock


\bibitem[Qiu and Lan(2024)]%
        {qiu2024interactive}
\bibfield{author}{\bibinfo{person}{Huachuan Qiu} {and} \bibinfo{person}{Zhenzhong Lan}.} \bibinfo{year}{2024}\natexlab{}.
\newblock \showarticletitle{Interactive agents: Simulating counselor-client psychological counseling via role-playing llm-to-llm interactions}.
\newblock \bibinfo{journal}{\emph{arXiv preprint arXiv:2408.15787}} (\bibinfo{year}{2024}).
\newblock


\bibitem[Qiu and Lan(2025)]%
        {PsyDial}
\bibfield{author}{\bibinfo{person}{Huachuan Qiu} {and} \bibinfo{person}{Zhenzhong Lan}.} \bibinfo{year}{2025}\natexlab{}.
\newblock \showarticletitle{PsyDial: {A} Large-scale Long-term Conversational Dataset for Mental Health Support}. In \bibinfo{booktitle}{\emph{Proceedings of the 63rd Annual Meeting of the Association for Computational Linguistics (Volume 1: Long Papers), {ACL} 2025, Vienna, Austria, July 27 - August 1, 2025}}. \bibinfo{publisher}{Association for Computational Linguistics}, \bibinfo{address}{Vienna, Austria}, \bibinfo{pages}{21624--21655}.
\newblock


\bibitem[Ribeiro et~al\mbox{.}(2013)]%
        {Psychology}
\bibfield{author}{\bibinfo{person}{Eug{\'e}nia Ribeiro}, \bibinfo{person}{Antonio~P Ribeiro}, \bibinfo{person}{Miguel~M Gon{\c{c}}alves}, \bibinfo{person}{Adam~O Horvath}, {and} \bibinfo{person}{William~B Stiles}.} \bibinfo{year}{2013}\natexlab{}.
\newblock \showarticletitle{How collaboration in therapy becomes therapeutic: The therapeutic collaboration coding system}.
\newblock \bibinfo{journal}{\emph{Psychology and Psychotherapy: Theory, Research and Practice}} \bibinfo{volume}{86}, \bibinfo{number}{3} (\bibinfo{year}{2013}), \bibinfo{pages}{294--314}.
\newblock


\bibitem[Sun et~al\mbox{.}(2025)]%
        {sun2025speed}
\bibfield{author}{\bibinfo{person}{Weigao Sun}, \bibinfo{person}{Jiaxi Hu}, \bibinfo{person}{Yucheng Zhou}, \bibinfo{person}{Jusen Du}, \bibinfo{person}{Disen Lan}, \bibinfo{person}{Kexin Wang}, \bibinfo{person}{Tong Zhu}, \bibinfo{person}{Xiaoye Qu}, \bibinfo{person}{Yu Zhang}, \bibinfo{person}{Xiaoyu Mo}, \bibinfo{person}{Daizong Liu}, \bibinfo{person}{Yuxuan Liang}, \bibinfo{person}{Wenliang Chen}, \bibinfo{person}{Guoqi Li}, {and} \bibinfo{person}{Yu Cheng}.} \bibinfo{year}{2025}\natexlab{}.
\newblock \showarticletitle{Speed Always Wins: {A} Survey on Efficient Architectures for Large Language Models}.
\newblock \bibinfo{journal}{\emph{CoRR}}  \bibinfo{volume}{abs/2508.09834} (\bibinfo{year}{2025}).
\newblock
\showeprint[arXiv]{2508.09834}


\bibitem[Team et~al\mbox{.}(2025)]%
        {team2025kimi}
\bibfield{author}{\bibinfo{person}{Kimi Team}, \bibinfo{person}{Angang Du}, \bibinfo{person}{Bohong Yin}, \bibinfo{person}{Bowei Xing}, \bibinfo{person}{Bowen Qu}, \bibinfo{person}{Bowen Wang}, \bibinfo{person}{Cheng Chen}, \bibinfo{person}{Chenlin Zhang}, \bibinfo{person}{Chenzhuang Du}, \bibinfo{person}{Chu Wei}, {et~al\mbox{.}}} \bibinfo{year}{2025}\natexlab{}.
\newblock \bibinfo{title}{Kimi-vl technical report}.
\newblock


\bibitem[Tran et~al\mbox{.}(2025)]%
        {Multi-Agent}
\bibfield{author}{\bibinfo{person}{Khanh{-}Tung Tran}, \bibinfo{person}{Dung Dao}, \bibinfo{person}{Minh{-}Duong Nguyen}, \bibinfo{person}{Quoc{-}Viet Pham}, \bibinfo{person}{Barry O'Sullivan}, {and} \bibinfo{person}{Hoang~D. Nguyen}.} \bibinfo{year}{2025}\natexlab{}.
\newblock \showarticletitle{Multi-Agent Collaboration Mechanisms: {A} Survey of LLMs}.
\newblock \bibinfo{journal}{\emph{CoRR}}  \bibinfo{volume}{abs/2501.06322} (\bibinfo{year}{2025}).
\newblock
\showeprint[arXiv]{2501.06322}


\bibitem[Wang et~al\mbox{.}(2025b)]%
        {AnnaAgent}
\bibfield{author}{\bibinfo{person}{Ming Wang}, \bibinfo{person}{Peidong Wang}, \bibinfo{person}{Lin Wu}, \bibinfo{person}{Xiaocui Yang}, \bibinfo{person}{Daling Wang}, \bibinfo{person}{Shi Feng}, \bibinfo{person}{Yuxin Chen}, \bibinfo{person}{Bixuan Wang}, {and} \bibinfo{person}{Yifei Zhang}.} \bibinfo{year}{2025}\natexlab{b}.
\newblock \showarticletitle{AnnaAgent: Dynamic Evolution Agent System with Multi-Session Memory for Realistic Seeker Simulation}. In \bibinfo{booktitle}{\emph{Findings of the Association for Computational Linguistics, {ACL} 2025, Vienna, Austria, July 27 - August 1, 2025}}. \bibinfo{publisher}{Association for Computational Linguistics}, \bibinfo{address}{Vienna, Austria}, \bibinfo{pages}{23221--23235}.
\newblock


\bibitem[Wang et~al\mbox{.}(2024)]%
        {MemoryMamba}
\bibfield{author}{\bibinfo{person}{Qianning Wang}, \bibinfo{person}{He Hu}, {and} \bibinfo{person}{Yucheng Zhou}.} \bibinfo{year}{2024}\natexlab{}.
\newblock \showarticletitle{MemoryMamba: Memory-Augmented State Space Model for Defect Recognition}.
\newblock \bibinfo{journal}{\emph{CoRR}}  \bibinfo{volume}{abs/2405.03673} (\bibinfo{year}{2024}).
\newblock
\showeprint[arXiv]{2405.03673}


\bibitem[Wang et~al\mbox{.}(2025a)]%
        {COMPEER}
\bibfield{author}{\bibinfo{person}{Yunxiao Wang}, \bibinfo{person}{Meng Liu}, \bibinfo{person}{Wenqi Liu}, \bibinfo{person}{Kaiyu Jiang}, \bibinfo{person}{Bin Wen}, \bibinfo{person}{Fan Yang}, \bibinfo{person}{Tingting Gao}, \bibinfo{person}{Guorui Zhou}, {and} \bibinfo{person}{Liqiang Nie}.} \bibinfo{year}{2025}\natexlab{a}.
\newblock \showarticletitle{{COMPEER:} Controllable Empathetic Reinforcement Reasoning for Emotional Support Conversation}.
\newblock \bibinfo{journal}{\emph{CoRR}}  \bibinfo{volume}{abs/2508.09521} (\bibinfo{year}{2025}).
\newblock
\showeprint[arXiv]{2508.09521}


\bibitem[Weizenbaum(1983)]%
        {ELIZA}
\bibfield{author}{\bibinfo{person}{Joseph Weizenbaum}.} \bibinfo{year}{1983}\natexlab{}.
\newblock \showarticletitle{{ELIZA} - {A} Computer Program For the Study of Natural Language Communication Between Man And Machine (Reprint)}.
\newblock \bibinfo{journal}{\emph{Commun. {ACM}}} \bibinfo{volume}{26}, \bibinfo{number}{1} (\bibinfo{year}{1983}), \bibinfo{pages}{23--28}.
\newblock


\bibitem[Wu et~al\mbox{.}(2022)]%
        {Anno-mi}
\bibfield{author}{\bibinfo{person}{Zixiu Wu}, \bibinfo{person}{Simone Balloccu}, \bibinfo{person}{Vivek Kumar}, \bibinfo{person}{Rim Helaoui}, \bibinfo{person}{Ehud Reiter}, \bibinfo{person}{Diego~Reforgiato Recupero}, {and} \bibinfo{person}{Daniele Riboni}.} \bibinfo{year}{2022}\natexlab{}.
\newblock \showarticletitle{Anno-mi: A dataset of expert-annotated counselling dialogues}. In \bibinfo{booktitle}{\emph{ICASSP 2022-2022 IEEE International Conference on Acoustics, Speech and Signal Processing (ICASSP)}}. IEEE, \bibinfo{publisher}{IEEE}, \bibinfo{address}{Singapore}, \bibinfo{pages}{6177--6181}.
\newblock


\bibitem[Xie et~al\mbox{.}(2025)]%
        {PsyDT}
\bibfield{author}{\bibinfo{person}{Haojie Xie}, \bibinfo{person}{Yirong Chen}, \bibinfo{person}{Xiaofen Xing}, \bibinfo{person}{Jingkai Lin}, {and} \bibinfo{person}{Xiangmin Xu}.} \bibinfo{year}{2025}\natexlab{}.
\newblock \showarticletitle{PsyDT: Using LLMs to Construct the Digital Twin of Psychological Counselor with Personalized Counseling Style for Psychological Counseling}. In \bibinfo{booktitle}{\emph{Proceedings of the 63rd Annual Meeting of the Association for Computational Linguistics (Volume 1: Long Papers), {ACL} 2025, Vienna, Austria, July 27 - August 1, 2025}}. \bibinfo{publisher}{Association for Computational Linguistics}, \bibinfo{address}{Vienna, Austria}, \bibinfo{pages}{1081--1115}.
\newblock


\bibitem[Xu et~al\mbox{.}(2025)]%
        {AutoCBT}
\bibfield{author}{\bibinfo{person}{Ancheng Xu}, \bibinfo{person}{Di Yang}, \bibinfo{person}{Renhao Li}, \bibinfo{person}{Jingwei Zhu}, \bibinfo{person}{Minghuan Tan}, \bibinfo{person}{Min Yang}, \bibinfo{person}{Wanxin Qiu}, \bibinfo{person}{Mingchen Ma}, \bibinfo{person}{Haihong Wu}, \bibinfo{person}{Bingyu Li}, \bibinfo{person}{Feng Sha}, \bibinfo{person}{Chengming Li}, \bibinfo{person}{Xiping Hu}, \bibinfo{person}{Qiang Qu}, \bibinfo{person}{Derek~F. Wong}, {and} \bibinfo{person}{Ruifeng Xu}.} \bibinfo{year}{2025}\natexlab{}.
\newblock \bibinfo{title}{AutoCBT: An Autonomous Multi-agent Framework for Cognitive Behavioral Therapy in Psychological Counseling}.
\newblock
\showeprint[arXiv]{2501.09426}


\bibitem[Yang et~al\mbox{.}(2025)]%
        {yang2025qwen3}
\bibfield{author}{\bibinfo{person}{An Yang}, \bibinfo{person}{Anfeng Li}, \bibinfo{person}{Baosong Yang}, \bibinfo{person}{Beichen Zhang}, \bibinfo{person}{Binyuan Hui}, \bibinfo{person}{Bo Zheng}, \bibinfo{person}{Bowen Yu}, \bibinfo{person}{Chang Gao}, \bibinfo{person}{Chengen Huang}, \bibinfo{person}{Chenxu Lv}, {et~al\mbox{.}}} \bibinfo{year}{2025}\natexlab{}.
\newblock \bibinfo{title}{Qwen3 technical report}.
\newblock


\bibitem[Yang et~al\mbox{.}(2024)]%
        {MentaLLaMA}
\bibfield{author}{\bibinfo{person}{Kailai Yang}, \bibinfo{person}{Tianlin Zhang}, \bibinfo{person}{Ziyan Kuang}, \bibinfo{person}{Qianqian Xie}, \bibinfo{person}{Jimin Huang}, {and} \bibinfo{person}{Sophia Ananiadou}.} \bibinfo{year}{2024}\natexlab{}.
\newblock \showarticletitle{MentaLLaMA: Interpretable Mental Health Analysis on Social Media with Large Language Models}. In \bibinfo{booktitle}{\emph{Proceedings of the {ACM} on Web Conference 2024, {WWW} 2024, Singapore, May 13-17, 2024}}. \bibinfo{publisher}{{ACM}}, \bibinfo{address}{Singapore}, \bibinfo{pages}{4489--4500}.
\newblock


\bibitem[Ye et~al\mbox{.}(2025)]%
        {SweetieChat}
\bibfield{author}{\bibinfo{person}{Jing Ye}, \bibinfo{person}{Lu Xiang}, \bibinfo{person}{Yaping Zhang}, {and} \bibinfo{person}{Chengqing Zong}.} \bibinfo{year}{2025}\natexlab{}.
\newblock \showarticletitle{SweetieChat: {A} Strategy-Enhanced Role-playing Framework for Diverse Scenarios Handling Emotional Support Agent}. In \bibinfo{booktitle}{\emph{Proceedings of the 31st International Conference on Computational Linguistics, {COLING} 2025, Abu Dhabi, UAE, January 19-24, 2025}}. \bibinfo{publisher}{Association for Computational Linguistics}, \bibinfo{address}{Abu Dhabi, UAE}, \bibinfo{pages}{4646--4669}.
\newblock


\bibitem[Yidianling(2015)]%
        {yidianling}
\bibfield{author}{\bibinfo{person}{Yidianling}.} \bibinfo{year}{2015}\natexlab{}.
\newblock \bibinfo{title}{Online Psychological Counseling Platform}.
\newblock
\urldef\tempurl%
\url{https://www.yidianling.com}
\showURL{%
\tempurl}
\newblock
\shownote{Accessed: 2025-10-02}.


\bibitem[Young et~al\mbox{.}(2024)]%
        {young2024yi}
\bibfield{author}{\bibinfo{person}{Alex Young}, \bibinfo{person}{Bei Chen}, \bibinfo{person}{Chao Li}, \bibinfo{person}{Chengen Huang}, \bibinfo{person}{Ge Zhang}, \bibinfo{person}{Guanwei Zhang}, \bibinfo{person}{Guoyin Wang}, \bibinfo{person}{Heng Li}, \bibinfo{person}{Jiangcheng Zhu}, \bibinfo{person}{Jianqun Chen}, {et~al\mbox{.}}} \bibinfo{year}{2024}\natexlab{}.
\newblock \bibinfo{title}{Yi: Open foundation models by 01. ai}.
\newblock


\bibitem[Zarbo et~al\mbox{.}(2016)]%
        {zarbo2016integrative}
\bibfield{author}{\bibinfo{person}{Cristina Zarbo}, \bibinfo{person}{Giorgio~A Tasca}, \bibinfo{person}{Francesco Cattafi}, {and} \bibinfo{person}{Angelo Compare}.} \bibinfo{year}{2016}\natexlab{}.
\newblock \bibinfo{title}{Integrative psychotherapy works}.
\newblock \bibinfo{numpages}{2021}~pages.
\newblock


\bibitem[Zhang et~al\mbox{.}(2024a)]%
        {CPsyCoun}
\bibfield{author}{\bibinfo{person}{Chenhao Zhang}, \bibinfo{person}{Renhao Li}, \bibinfo{person}{Minghuan Tan}, \bibinfo{person}{Min Yang}, \bibinfo{person}{Jingwei Zhu}, \bibinfo{person}{Di Yang}, \bibinfo{person}{Jiahao Zhao}, \bibinfo{person}{Guancheng Ye}, \bibinfo{person}{Chengming Li}, {and} \bibinfo{person}{Xiping Hu}.} \bibinfo{year}{2024}\natexlab{a}.
\newblock \showarticletitle{CPsyCoun: {A} Report-based Multi-turn Dialogue Reconstruction and Evaluation Framework for Chinese Psychological Counseling}. In \bibinfo{booktitle}{\emph{Findings of the Association for Computational Linguistics, {ACL} 2024, Bangkok, Thailand and virtual meeting, August 11-16, 2024}}. \bibinfo{publisher}{Association for Computational Linguistics}, \bibinfo{address}{Vienna, Austria}, \bibinfo{pages}{13947--13966}.
\newblock


\bibitem[Zhang et~al\mbox{.}(2025)]%
        {Preference}
\bibfield{author}{\bibinfo{person}{Mian Zhang}, \bibinfo{person}{Shaun~M. Eack}, {and} \bibinfo{person}{Zhiyu~Zoey Chen}.} \bibinfo{year}{2025}\natexlab{}.
\newblock \bibinfo{title}{Preference Learning Unlocks LLMs' Psycho-Counseling Skills}.
\newblock
\showeprint[arXiv]{2502.19731}


\bibitem[Zhang et~al\mbox{.}(2024b)]%
        {ESCoT}
\bibfield{author}{\bibinfo{person}{Tenggan Zhang}, \bibinfo{person}{Xinjie Zhang}, \bibinfo{person}{Jinming Zhao}, \bibinfo{person}{Li Zhou}, {and} \bibinfo{person}{Qin Jin}.} \bibinfo{year}{2024}\natexlab{b}.
\newblock \showarticletitle{ESCoT: Towards Interpretable Emotional Support Dialogue Systems}. In \bibinfo{booktitle}{\emph{Proceedings of the 62nd Annual Meeting of the Association for Computational Linguistics (Volume 1: Long Papers), {ACL} 2024, Bangkok, Thailand, August 11-16, 2024}}. \bibinfo{address}{Vienna, Austria}, \bibinfo{pages}{13395--13412}.
\newblock


\bibitem[Zhou et~al\mbox{.}(2023)]%
        {zhou2023thread}
\bibfield{author}{\bibinfo{person}{Yucheng Zhou}, \bibinfo{person}{Xiubo Geng}, \bibinfo{person}{Tao Shen}, \bibinfo{person}{Chongyang Tao}, \bibinfo{person}{Guodong Long}, \bibinfo{person}{Jian{-}Guang Lou}, {and} \bibinfo{person}{Jianbing Shen}.} \bibinfo{year}{2023}\natexlab{}.
\newblock \showarticletitle{Thread of Thought Unraveling Chaotic Contexts}.
\newblock \bibinfo{journal}{\emph{CoRR}}  \bibinfo{volume}{abs/2311.08734} (\bibinfo{year}{2023}).
\newblock
\showeprint[arXiv]{2311.08734}


\bibitem[Zhou et~al\mbox{.}(2024)]%
        {zhou2024visual}
\bibfield{author}{\bibinfo{person}{Yucheng Zhou}, \bibinfo{person}{Xiang Li}, \bibinfo{person}{Qianning Wang}, {and} \bibinfo{person}{Jianbing Shen}.} \bibinfo{year}{2024}\natexlab{}.
\newblock \showarticletitle{Visual In-Context Learning for Large Vision-Language Models}. In \bibinfo{booktitle}{\emph{Findings of the Association for Computational Linguistics, {ACL} 2024, Bangkok, Thailand and virtual meeting, August 11-16, 2024}}. \bibinfo{publisher}{Association for Computational Linguistics}, \bibinfo{pages}{15890--15902}.
\newblock


\bibitem[Zhou et~al\mbox{.}(2025)]%
        {zhou2025mam}
\bibfield{author}{\bibinfo{person}{Yucheng Zhou}, \bibinfo{person}{Lingran Song}, {and} \bibinfo{person}{Jianbing Shen}.} \bibinfo{year}{2025}\natexlab{}.
\newblock \showarticletitle{{MAM:} Modular Multi-Agent Framework for Multi-Modal Medical Diagnosis via Role-Specialized Collaboration}. In \bibinfo{booktitle}{\emph{Findings of the Association for Computational Linguistics, {ACL} 2025, Vienna, Austria, July 27 - August 1, 2025}}. \bibinfo{publisher}{Association for Computational Linguistics}, \bibinfo{pages}{25319--25333}.
\newblock


\bibitem[Zhu et~al\mbox{.}(2025)]%
        {Arena}
\bibfield{author}{\bibinfo{person}{Shijing Zhu}, \bibinfo{person}{Zhuang Chen}, \bibinfo{person}{Guanqun Bi}, \bibinfo{person}{Binghang Li}, \bibinfo{person}{Yaxi Deng}, \bibinfo{person}{Dazhen Wan}, \bibinfo{person}{Libiao Peng}, \bibinfo{person}{Xiyao Xiao}, \bibinfo{person}{Rongsheng Zhang}, \bibinfo{person}{Tangjie Lv}, \bibinfo{person}{Zhipeng Hu}, \bibinfo{person}{Fangfang Li}, {and} \bibinfo{person}{Minlie Huang}.} \bibinfo{year}{2025}\natexlab{}.
\newblock \bibinfo{title}{{\(\Psi\)}-Arena: Interactive Assessment and Optimization of LLM-based Psychological Counselors with Tripartite Feedback}.
\newblock
\showeprint[arXiv]{2505.03293}


\end{thebibliography}
